\documentclass{article}

\pdfoutput=1

\usepackage[numbers]{natbib}
\usepackage[final]{neurips_2022}

\usepackage{amsmath}
\usepackage{amssymb}
\usepackage{mathtools}
\usepackage{amsthm}
\usepackage{graphicx}

\usepackage{algorithmicx}
\usepackage{algorithm}%
\usepackage{algpseudocode}%

\usepackage[utf8]{inputenc} %
\usepackage[T1]{fontenc}    %
\usepackage{url}            %
\usepackage{booktabs}       %
\usepackage{amsfonts}       %
\usepackage{nicefrac}       %
\usepackage{microtype}      %
\usepackage{xcolor}         %
\usepackage{comment}

\usepackage{caption}
\usepackage{subcaption}
\usepackage{wrapfig}
\usepackage{bm}

\usepackage{hyperref}       %

\newcommand{\algo}{Adaptive Distillation of Adapters}
\newcommand{\al}{ADA}

\theoremstyle{plain}

\theoremstyle{definition}

\theoremstyle{remark}

\title{Memory Efficient Continual Learning with Transformers}

\author{%
  Beyza Ermis \\
  Amazon Web Services \\
  \texttt{ermibeyz@amazon.com} \\
  \And
  Giovanni Zappella \\
  Amazon Web Services \\
  \texttt{zappella@amazon.com} \\
  \And
  Martin Wistuba \\
  Amazon Web Services \\
  \texttt{marwistu@amazon.com} \\
  \And
  Aditya Rawal \\
  Amazon Web Services \\
  \texttt{marwistu@amazon.com} \\
  \And
  C\'edric Archambeau \\
  Amazon Web Services \\
  \texttt{cedrica@amazon.com} \\  
}

\begin{document}

\maketitle

\begin{abstract}

In many real-world scenarios, data to train machine learning models becomes available over time. 
Unfortunately, these models struggle to continually learn new concepts without forgetting what has been 
learnt in the past. This phenomenon is known as catastrophic forgetting and it is difficult to prevent due to 
practical constraints. For instance, the amount of data that can be stored or the computational resources 
that can be used might be limited. Moreover, applications increasingly rely on large pre-trained neural 
networks, such as pre-trained Transformers, since the resources or data might not be available in 
sufficiently large quantities to practitioners to train the model from scratch. In this paper, we devise a 
method to incrementally train a model on a sequence of tasks using pre-trained Transformers and extending 
them with Adapters. Different than the existing approaches, our method is able to scale to a large number of 
tasks without significant overhead and allows sharing information across tasks. On both image and text 
classification tasks, we empirically demonstrate that our method maintains a good predictive performance 
without retraining the model or increasing the number of model parameters over time. The resulting model is
also significantly faster at inference time compared to Adapter-based state-of-the-art methods.

\end{abstract}

\section{Introduction}
\label{sec:intro}

Transformers~\cite{vaswani2017attention}, e.g. BERT~\cite{kenton2019bert}, have shown their effectiveness in various natural language 
processing (NLP) tasks such as classification~\cite{ke2020continual}, Natural Language Inference~\cite{pfeiffer2020mad,pfeiffer-etal-2021-adapterfusion}, 
and Question Answering~\cite{greco2019psycholinguistics}. 
Inspired by this achievement, some pioneering works have recently been introduced on adapting Transformers architectures to Computer Vision (CV).
Vision Transformers~\cite{dosovitskiy2020image,touvron2021training} showed that a pure Transformer applied directly to a sequence of image patches 
can perform well on image classification tasks. Besides, some recent studies~\cite{yin2019benchmarking,brown2020language,li2022technical} showed that
Transformers are generalized to new domains given only a few samples. 
Transformers show a great ability to learn complex concepts but when confronted with a sequence of different tasks
they tend to ``overwrite'' the previously learnt concepts. In general, deep networks suffer heavily from this 
phenomenon, called \textit{catastrophic forgetting} (CF)~\cite{mccloskey1989catastrophic}, impeding continual or lifelong learning.
In the last few years, a growing body of works attempted to tackle CF
in continual learning (CL) \cite{douillard2020podnet,hou2019learning,kirkpatrick2017overcoming,rebuffi2017icarl,wu2019large,yan2021dynamically} but
most of them are not able to meet the scale or accuracy requirements of real-world applications. 
Moreover, adapting large-scale pre-trained Transformer models to downstream tasks via fine-tuning is the method of choice
in NLP applications, posing the need for methods that can directly work with pre-trained models instead of requiring the training of
a new model from scratch~\cite{douillard2021dytox}.

In this work, we tackle both text and image classification problems in a setting where the number of tags or classes associated to the 
input data grows over time. In fact, the ability to continually extend the set of tags or classes used to categorize the content is a major
problem in many applications.
For example, newspapers can tag news according to topics of interest such as ``sport'', ``politics'', ``food''
by using a pre-trained language model and refining it using a few hundred pre-tagged articles.
New tags may appear over time, for example ``COVID-19'' was a completely unknown news category in 2019 but appeared frequently since it is emerged.
In these cases, retraining models from scratch is often impractical and can lead to 
inconsistencies in the labeling when compared to the one provided by the previous model. 
In particular, we focus on incrementally extending classifiers based on pre-trained Transformer models given
their ubiquity in NLP and the growing interest in CV.

To address the issue of incremental fine-tuning of pre-trained Transformers in the sequential learning setting without CF,
we propose \algo{} (\al{}). \al{} leverages Adapters~\cite{houlsby2019parameter}, a specialized neural network module that adds new parameters
to the neural network and a distillation mechanism to consolidate the new information with the previously learnt knowledge in a fixed amount of 
parameters with little amount of forgetting.
This method allows the user to control the memory consumption, while retaining state-of-the-art performance when running the algorithm on sequences of tens of tasks.
This tight memory control is important in industrial applications.
The alternative, a model growing in size with the number of tasks, would require a change in hardware to adapt to the growing memory requirements of the deployed model.
This would be problematic since a practitioner will incur into higher risk of system instability and 
be forced to make conservative hardware choices.

The main contribution of our work is \al{}, an algorithm that can achieve high predictive performance on 
both text and image classification in different continual learning scenarios.
\al{} also provides lower inference time and uses an order of magnitude fewer parameters than state of the art methods such AdaptersFusion~\cite{pfeiffer-etal-2021-adapterfusion}.
Additionally, we implemented Adapters for vision Transformers and empirically demonstrated their effectiveness.

\section{Related Work}
\label{sec:related_work}

Adapters~\cite{houlsby2019parameter} were proposed for fine-tuning of pre-trained language models and were studied for the multi-task
setting. AdapterFusion~\cite{pfeiffer-etal-2021-adapterfusion} provides state-of-the-art performance by composing the pre-trained Adapters
and it can simply be repurposed for preventing CF in CL by learning one Adapter for every new task. 
While it has been shown that the number of additional model parameters per Adapter is significantly smaller than the number of parameters used in the pre-trained
model~\cite{houlsby2019parameter} (e.g., $3.6\%$  of the parameters of the pre-trained model), since both Adapters and AdapterFusion require to store all the model parameters,
the memory consumption increases rapidly with the number of tasks. 
In the case of a model being trained on 30 tasks, we would have to add more parameters than the number of pre-trained Transformer parameters (details in Section~\ref{sec:experiments}).
The linear increase in memory and storage consumption making the method unsuitable for CL.

Recent work studied catastrophic forgetting~\citep{sun2019lamol,chuang2020lifelong,ke2020continual,liu2020exploring} 
and incremental learning~\cite{xia2021incremental} for NLP and CV~\cite{li2022technical,douillard2021dytox}.
Pasanuru et al.~\cite{pasunuru2021continual} focus on the few-shot setting where only a few data points are available for each task. 
Ke et al.~\cite{ke2021adapting} proposed an architecture to achieve both CF prevention and knowledge transfer. This method has some similarity
to AdapterBERT~\cite{houlsby2019parameter} since they insert a CL plug-in module in two locations in BERT.
A CL-plugin is a capsule network~\cite{sabour2017dynamic} that uses one separate
capsule~\cite{hinton2011transforming} (2-layer fully connected network) for each task, and like Adapters, memory increases linearly over the time. 
In addition, this algorithm requires to learn task masks to address knowledge transfer, which is costly to compute. 
Among those recent works, only a few~\cite{li2022technical,douillard2021dytox} have applied the Transformers architecture to CL on image datasets.
In~\cite{li2022technical}, for each new task, the model is copied and fixed to be used as the teacher model in the distillation phase. 
The student model is trained on both new task samples together with the knowledge distillation loss that uses samples from old tasks which is stored
in the rehearsal memory. In~\cite{douillard2021dytox}, the authors aim to learn a unified model that will classify an increasingly
growing number of classes by building upon a new architecture. However, they need to train a new Transformer, 
where the process is very costly and contrast with our goal of using public pre-trained models.
To the best of our knowledge there is no method able to leverage public pre-trained Transformers while keeping the number
of model parameters constant and retaining state-of-the-art predictive performance.

\section{Problem Setup and Preliminaries}
\label{sec:setup}

\paragraph{Problem Setup.} A sequence of classification tasks $\lbrace T_1, \dots, T_N \rbrace$ are given where each task 
$T_i$ contains a different set of data sample (text or image)-label training pairs $(x_{1:t}^i, y_{1:t}^i)$ and 
contains $c$ new classes namely $Y_i=\lbrace Y_i^1, \dots, Y_i^c\rbrace$ with $t$ examples for each new class.
The goal of the learner is to learn a set of parameters $\tilde{\Theta}$ such that 
$\frac{1}{N} \sum_{i \in \{1,\dots, N\}} \text{loss}(T_i; \tilde{\Theta})$ is minimized.
The task identifier is provided to the learner with every new batch of data.
Moreover, in our specific case, $\tilde{\Theta}$ is composed of a set of parameters $\Theta$
provided by a pre-trained model and, depending on the algorithm, some additional parameters
which need to be learned for each specific task.
In its simplest case, this additional set of model parameters can just be a head model,
but some algorithms use significantly more elaborate functions.
In the case of the labeling application described in Section~\ref{sec:intro}, each task represents
a label and the learner creates a new binary classifier for each label.

For the training of task $T_i$, the learner can only access the newly added examples and label names in this task. 
To evaluate the learner, the test data consists of examples across all the previous tasks, where the potential label
space for the test example is $Y_1^{1:c} \cup Y_2^{1:c} \cup \dots \cup Y_N^{1:c}$.
All methods that we define in the following sections receive as input a pre-trained model $f_{\Theta}(.)$,
e.g., BERT~\cite{kenton2019bert}, that is able to extract high quality representations from the input data.

\paragraph{Adapters.} Adapters were proposed by~\cite{houlsby2019parameter} as an alternative to fine-tuning in NLP. 
They add new modules between layers of a pre-trained network called \emph{Adapters}. These modules are feed-forward
layers that project the original feature size to a smaller dimension and projects them to the original size thereafter,
ensuring that the number of parameters stays substantially small as compared to the original model.
(See Appendix~\ref{app:vis_adptr} for the details of the Adapter architectures.)
Adapters share the pre-trained model parameters $\Theta$ across all tasks and introduce a small number of task-specific
parameters $\Phi_i$ without affecting previous ones. The model is initialized with parameters of a pre-trained model $\Theta$.
For each of the task $i \in \lbrace 1, \dots, N \rbrace$ where $N$ is the total number of tasks, a set of new and randomly
initialized Adapter parameters $\Phi_i$ are introduced.
The parameters $\Theta$ are fixed and only the parameters $\Phi_i$ are trained when a new task is added. 
This makes it possible to train Adapters for all $N$ tasks, and store the corresponding knowledge in designated parts of the model. 
The objective for each task $i \in \lbrace 1, \dots, N \rbrace$ is of the form:
$\Phi_i \leftarrow \arg\min_\Phi L_i(D_i;\Theta,\Phi)$. 

AdapterFusion~\cite{pfeiffer-etal-2021-adapterfusion}, has been proposed to mitigate
the lack of knowledge sharing across tasks. It works in two phases: i) in the knowledge extraction stage, 
adapters, which encapsulate the task-specific information, are 
learnt for each of the $N$ tasks; while ii) in the knowledge composition stage,
the set of $N$ Adapters are combined by using additional parameters $\Psi$. 
The additional parameters $\Psi_i$ for task $i$ are defined as:
$\Psi_i \leftarrow \arg\min_\Psi L_i(D_i; \Theta, \Phi_1, \dots, \Phi_i, \Psi)$. 
While this provides good predictive performance, in the CL setting, new tasks are added 
sequentially and storing a large set of Adapters $\Phi_1, \dots, \Phi_N$ is practically infeasible.

\section{\algo{} (\al{})}
\label{sec:ada}

To address the issues we mentioned in the previous sections, we propose \algo{} (\al{}). 
\al{} keeps a fixed amount of Adapters in memory and takes transferability of representations into account to effectively consolidate newly
created Adapters with previously created ones. %
\al{} works in two steps: i) it trains a new  Adapter and classification head, 
which we refer as the \emph{new} model, using the training dataset of the new task; 
ii) it consolidates the \emph{old} model with the new model.
To better control the memory usage, \al{} has a fix budget for the number of 
Adapters $K$ that are stored in a pool of old models. 
In the consolidation phase, the algorithm selects one of the models in the pool 
using scores that quantify the transferable information contained in the representations
they provide. In the following sections, we explain the components of \al{} and how they work.

\subsection{Distillation of Adapters}
\label{sec:distillation}

For each new task $T_n$, the Adapter parameters $\Phi_n$ are added to the model,
while the pre-trained model parameters $\Theta$ are kept frozen and are never changed.
Only the task-specific Adapter parameters $\Phi_n$ and the head model parameters $h_n$
are trained for the current task. The model $f_n(x;\Theta,\Phi_n, h_n)$, 
with parameters $\Theta$, $\Phi_n$  and $h_n$ is called the \emph{new} model. 
The head model parameters are fixed after training the new model and they are not updated
during model consolidation. When a prediction for a task $T_i$ is required, the corresponding Adapter 
$\Phi_{\gamma(i)}$ and head model $h_i$ is called. $\gamma$ is a mapping from the task id to the 
corresponding Adapter in the pool or to the newly trained Adapter.
We abuse notation defining $f$ as the function returning the output (\emph{logits}) on all tasks:
\begin{equation}
\label{eq:flogits}
f(x;\Theta,\Phi, h) = \left[ f(x;\Theta,\Phi_{\gamma(1)},h_1), \dots, f(x;\Theta,\Phi_{\gamma(n-1)},h_{n-1}), f(x;\Theta,\Phi_{\gamma(n)},h_n) \right]
\end{equation}

For the consolidation step, an Adapter from the pool $\bm{\Phi}$ is selected as explained in Section~\ref{sec:distillation}
and new collection of Adapter pool $\bm{\Phi}'$ is created where the selected Adapter is replaced with $\Phi_c$.
$\Phi_c$ denotes the consolidated model parameters and preliminarily the parameters are randomly initialized. 
Similarly, a copy of $\gamma$ is created to map the old tasks that are associated to the selected Adapter and the new task $n$ to $\Phi_c$.
The consolidation then has the following objective:
\begin{equation}
\label{eq:consolidation}
\min_{\Phi_c} \frac{1}{\vert \mathcal{D}_{distill} \vert}\sum_{i=1}^{\vert \mathcal{D}_{distill} \vert} (f(x_i; \Theta, \bm{\Phi}, h)-f(x_i; \Theta, \bm{\Phi}', h))^2
\end{equation}
where $\mathcal{D}_{distill}$ denotes the unlabeled training data used for distillation, and the distillation loss is computed as the difference
between the logits produced by the existing specialist models denoted by $f(x_i; \Theta, \bm{\Phi}, h)$ and the consolidated model denoted by 
$f(x_i; \Theta, \bm{\Phi}', h)$ based on $L_2$ loss.
After $\Phi_c$ has been trained, $\bm{\Phi}$ is swapped with $\bm{\Phi}'$.
This is a high-level view of the mechanism, our implementation is optimized to avoid copying models when not necessary.

This procedure follows the double distillation loss~\cite{zhang2020class} which is originally proposed for class incremental learning
to train a new Adapter that is used with the pre-trained model to classify both old and new tasks. The main idea is first training a
separate model for the new class(es) using labeled data, and then combining the new and old models using unlabeled distillation data
via a double distillation training objective. We generalize this solution to our case where we have a set of teacher models kept in Adapter
pool $\bm{\Phi}$ and train a student model $\Phi_c$. Double distillation procedure and the alternative solutions for distillation
are discussed in Appendix~\ref{app:related} but this solution was the best performing one in our experiments.

While several different data sources can be used to populate the buffer, such as using auxiliary external data~\cite{zhang2020class}
or generating synthetic data~\cite{chawla2021data}, in this work we populate the buffer using covariates from previous tasks selected
with Reservoir Sampling~\cite{vitter1985random}. This simple mechanism may not be the most effective, 
but it will guarantee that no advantage is given to \al{} in the experimental comparison.

\subsection{Adapter Selection for Distillation}
\label{sec:distillation}

In the previous section, we assumed the Adapter to be consolidated as given but \al{} keeps a pool of Adapters and
the selection of the Adapter to be distilled is an important part of the algorithm.
In fact, our empirical observations show that a random selection of the Adapter
provides poor performance (see Section~\ref{sec:ablation}). 
The intuition behind our selection mechanism is the following: 
since a specialized head for every task is created, we can assume that when the features provided by the associated Adapter 
are highly informative, the updates (i.e., the gradients applied) will be small. 
At the same time, training a new head with every Adapter in the pool in order to
observe which one is the most effective would increase the amount of computation required
and significantly impact the usability of the method. The problem of computing the information
carried by a representation in an efficient manner has been already studied in the transfer learning
community~\cite{bao2019information,tran2019transferability,tan2021otce}.

While, the aim of that research is completely different and, to the best of our knowledge, 
there is no clear relation between transferability and forgetting, the mathematical foundation 
of this work are closely related to our intuition. 
In fact, scores like TransRate~\cite{huang2021frustratingly} employ the mutual information between the
features provided by a pre-trained model and the target labels for the task at hand. 
When the mutual information is high, the transferability is high. 
More specifically, the knowledge transfer from a source task $T_s$ to a target task $T_t$ is measured as:
\begin{align}
\text{TrR}_{T_s\rightarrow T_t}(f(\Theta,\Phi_{\gamma(s)})) = H(Z) - H(Z|Y) ,
\label{eq:transrate}
\end{align}
where $Y$ are the labels of target examples and $Z=f(X;\Theta,\Phi_{\gamma(s)})$ are features of them extracted by the pre-trained model and the
Adapter associated to the source task.

TransRate is not the only score designed to quantify transferability between a pre-trained
model and a new task: \emph{Log Expected Empirical Prediction} (LEEP)~\cite{nguyen2020leep} 
is a well-known alternative. Also in this case, the score was designed with a different application
in mind, but it leverages the conditional distribution of the target label
given the source label to quantify the how informative the information provided by the source model is.
Specifically, LEEP is a three steps method. 
At Step 1, it computes dummy label distributions of the inputs $f(X; \Theta, \Phi_{\gamma(t)}, h_t)$ in the target data set $\mathcal{D}$. 
At Step 2, it computes the empirical conditional distribution $\hat{P}(y|z)$ of target label $y$ given the source label $z$. 
At Step 3, it computes LEEP using $f(X; \Theta, \Phi_{\gamma(s)}, h_s)$ and $\hat{P}(y|z)$:  
\small
\begin{align}
L(f(\Theta,\Phi_{\gamma(s)}, h_s), \mathcal{D}) = \frac{1}{m}\sum_{i=1}^m\log\left(\sum_{z\in\mathcal{Z}}\hat{P}(y|z) f(X; \Theta, \Phi_{\gamma(s)}, h_s)_z \right) ,
\label{eq:leep}
\end{align}
\normalsize
where $z$ is a dummy label randomly drawn from $f(X; \Theta, \Phi_{\gamma(s)}, h_s)$ and $y$ is randomly drawn from $\hat{P}(y|z)$.
We selected TransRate and LEEP for their simplicity and their ability to provide a quantification
without training but practitioners can replace these scores with different ones as they see fit.

\subsection{Algorithm}
\label{sec:ada_algo}

\begin{wrapfigure}{L}{0.54\textwidth}
\scalebox{0.86}{
\begin{minipage}{0.60\textwidth}
\begin{algorithm}[H]
\small
\caption{Adaptive Distillation of Adapters (ADA)}
\begin{algorithmic}%
\Require $\Theta$: pre-trained model, $K$: adapters pool size
\State Freeze $\Theta$ and create $\gamma = Map()$
\For{$n \gets 1$ to $N$}
\State A task $T_n$ is received
\State Initialize $\Phi_n$
\State Process $T_n$, train new model $f_n(x;\Theta,\Phi_n, h_n)$ 
\State Sample from $T_n$ and add to distillation data $\mathcal{D}_{distill}$
\If {$n \leq K$}
\State Store $f_n$ in the pool
\Else
\State $j^{\ast} \gets \arg\max_{j \in \{1, \dots, K\}} \operatorname{TranScore}(T_n,f_j)$
\State Add ($n, j^{\ast}$) to $\gamma$
\State Consolidate model: \\
\hspace*{13mm} $f_{j^{\ast}}$ = $\operatorname{Distill}(f_{j^{\ast}}, f_n, \mathcal{D}_{distill})$
\EndIf
\State Serve predictions for any task $i \leq n$ using $f_{\gamma(i)}$
\EndFor
\vspace*{1mm}
\State $\operatorname{Distill}(f_i, f_j, \mathcal{D}_{distill})$:
\State \hspace*{2mm} Get soft targets $\hat{y}_i$ from old model $f_i$ with $\mathcal{D}_{distill}$
\State \hspace*{2mm} Get soft targets $\hat{y}_j$ from new model $f_j$ with $\mathcal{D}_{distill}$
\State \hspace*{2mm} Initialize $\Phi_c$
\State \hspace*{2mm} Compute distillation loss and train model $f(x;\Theta,\Phi_c)$ as defined in Eq.~\ref{eq:consolidation}
\State \hspace*{2mm} \textbf{return} $f$
\end{algorithmic}
\label{algo:ADA}
\end{algorithm}
\end{minipage}
}
\end{wrapfigure}
\normalsize

\al{} is detailed in Algorithm~\ref{algo:ADA}. The graphical workflow of the algorithm is shown in Appendix~\ref{fig:ADAalg}.
For every new task, the algorithm trains a new adapter and head model (called $\Phi_n$ and $h_n$).
If the adapters pool did not reach the maximum size yet (controlled by $K$), it just adds it to the pool.
If the pool reached the maximum size, the algorithm is forced to select one of the adapters already
in the pool and distill it together with the newly trained one.
In order to select which adapter to distill, \al{} uses the transferability scores (e.g., LEEP or TransRate).
Once the adapter in the pool with the highest transferability score (called $f_{j^{\ast}}$) is identified,
it consolidates that adapter and the newly trained one into a new adapter and replaces the old one present in the pool.
In order to be able to make effective predictions, the algorithm also keeps a mapping $\gamma$ of which
adapter in the pool must be used in combination with each of the task-specific heads. 
\linebreak

\section{Experiments}
\label{sec:experiments}
In this section, we empirically validate our adapter distillation approach on text and image classification tasks
and show that \al{} achieves similar performance to AdapterFusion while consuming significantly less memory.
We dedicate Section~\ref{sec:ablation} to ablation studies providing further insights into the mechanisms implemented in \al{} 
and their contribution.

\textbf{Datasets and experimental setup.} 
We use three text datasets for multi-label text classification: Arxiv Papers~\cite{yang2018sgm} (paper classification), 
Reuters (RCV1-V2)~\cite{lewis2004rcv1} (news classification), Wiki-30K~\cite{zubiaga2012enhancing} (Wikipedia article classification)
and two dataset for image classification: CIFAR100~\cite{krizhevsky2009learning}
and MiniImageNet~\cite{russakovsky2015imagenet}. Details about the datasets are given in Appendix~\ref{app:datasets}.

For the \emph{multi-label text classification} experiments, we first sample a sequence of labels from the label space.
Then, we create a balanced binary classification task for each label by sampling the same amount of positive data 
points from the label considered and negative data points from the labels preceding the current one in the sequence.
After splitting the data in training and test set, we provide the algorithm with the 
training set and subsequently measure its performance on the test set.
The algorithm never observes any data point in the test set and, more generally, every
data point in the dataset is used only once.
For Arxiv Papers and Reuters datasets, we created 20 tasks and for Wiki-30K 60. We fixed the number of training samples 
per task to 100. The test set consists of 40 data points on Reuters and of 100 data points on Arxiv and Wiki-30K.

For \emph{image classification}, we design two scenarios. In the first scenario, each new task is a balanced 
binary classification problem. Each class can be selected to be the positive class only once.
In the second scenario each task is a balanced multi-class classification problem with 5 classes. 
In both cases we provide the learner with 50 data points per class both at training and test time:
in the first scenario each task will have a training set of 250 data points and in the second case
of 100 data points. The total number of tasks is fixed to 20 for both scenarios.
The distillation memory size is fixed to $1000$ for Wiki-30K which has a larger number of tasks, and to $500$ for the others.

\textbf{Metrics.} In~\cite{lopez2017gradient}, three metrics that we discuss in the following are defined to evaluate the performance of a CL method.
We use these metrics to evaluate our methods. It is considered that we have access to a test set for each of the $N$ tasks in $\lbrace T_1, \dots, T_N\rbrace$.
After the model finishes learning about the task $T_i$, its test performance are evaluated on all $N$ tasks. By doing so, a matrix $R \in \mathbb{R}^{N\times N}$
is constructed where $R_{i,j}$ is the test classification accuracy of the model on task $T_j$ after observing the last sample from task $T_i$.
Letting $\bar{b}$ be the vector of test accuracies for each task at random initialization, the three metrics are defined: 
i) Average Accuracy = $\frac{1}{N} \sum_{i=1}^N R_{N,i}$, ii) Backward Transfer (BWT) = $\frac{1}{N-1} \sum_{i=1}^{N-1} R_{N,i} - R_{i,i}$
and iii) Forward Transfer (FWT) = $\frac{1}{N-1} \sum_{i=2}^N R_{i-1,i} - \bar{b}_i$. (The larger these metrics, the better the model.)
All the results in this section are averaged over 5 runs.

\textbf{Baselines.} We compare \al{} the following baselines.
1) \emph{Fine-tuning head model (B1)}: We freeze the pre-trained representation and only fine-tune the output layer of each classification task. 
The output layer is multiple-head binary classifier that we also use for the other methods.
2) \emph{Fine tuning the full model (B2)}: We fine-tune both the pre-trained representation and the output layer for each classification task.  
3) \emph{Adapters}~\cite{houlsby2019parameter}: We train and keep separate Adapters for each classification task as well as the head models. 
4) \emph{AdapterFusion}~\cite{pfeiffer-etal-2021-adapterfusion}: It is a two stage learning algorithm that leverages knowledge from multiple tasks
by combining the representations from several task Adapters in order to improve the performance on the target task. 
This follows exactly the solution depicted in Section~\ref{sec:setup}.
5) \emph{Experience Replay (ER)}~\cite{rolnick2018experience}: ER is a commonly used baseline in Continual Learning that stores a subset of data for each task 
and then ``replays'' the old data together with the new one to avoid forgetting old concepts.
\cite{d2019episodic} propose to use such a memory module for sparse experience replay and local adaptation in the language domain.
This method stores all training examples, in order to achieve optimal performance. 
To make this method comparable with \emph{adapter-based} methods, we freeze pre-trained 
representation, add a single adapter parameters $\Phi$ and train the adapter by replaying 
examples from old tasks while training using data from the new task.
In order to keep baselines comparable we assign to ER the same amount of memory is used for the distillation buffer in \al{}.
In addition to these baselines, we use one special case of \al{} with K=1 as a baseline to demonstrate the advantage of effective consolidation of Adapters.
\begin{figure}[ht!]
\begin{minipage}{0.33\textwidth}
\centering
\includegraphics[width=\textwidth] {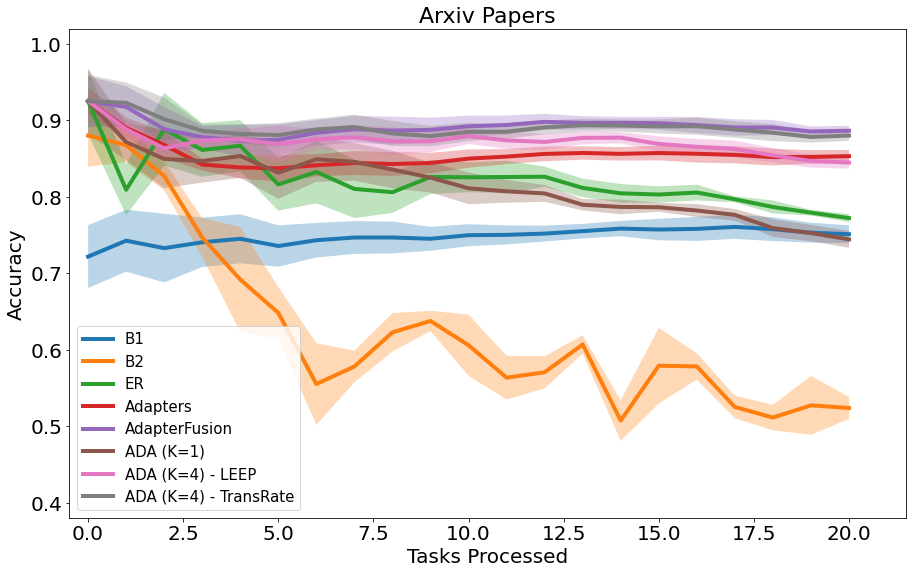}
\end{minipage}
\begin{minipage}{0.33\textwidth}
\centering
\includegraphics[width=\textwidth] {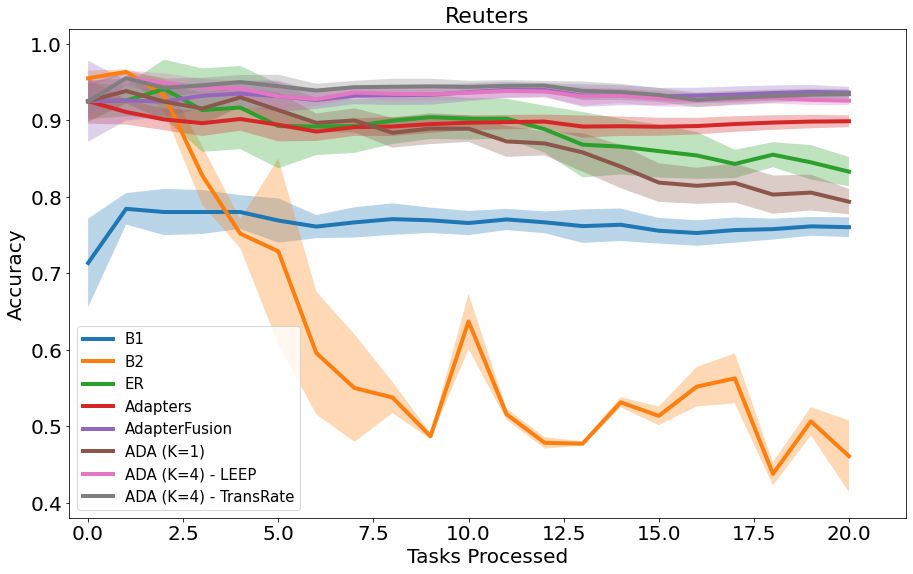}
\end{minipage}
\begin{minipage}{0.33\textwidth}
\centering
\includegraphics[width=\textwidth] {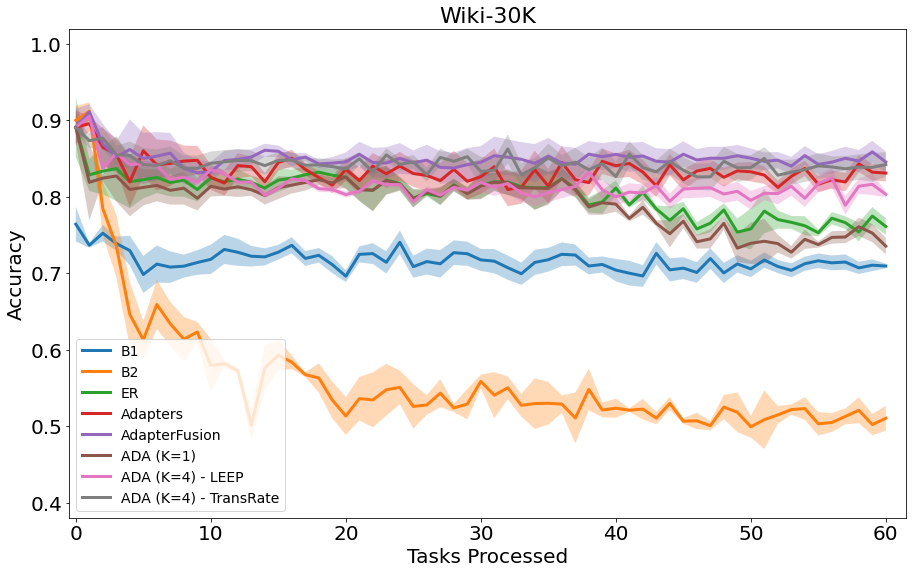}
\end{minipage}
\\
\begin{minipage}{0.33\textwidth}
\centering
\includegraphics[width=\textwidth] {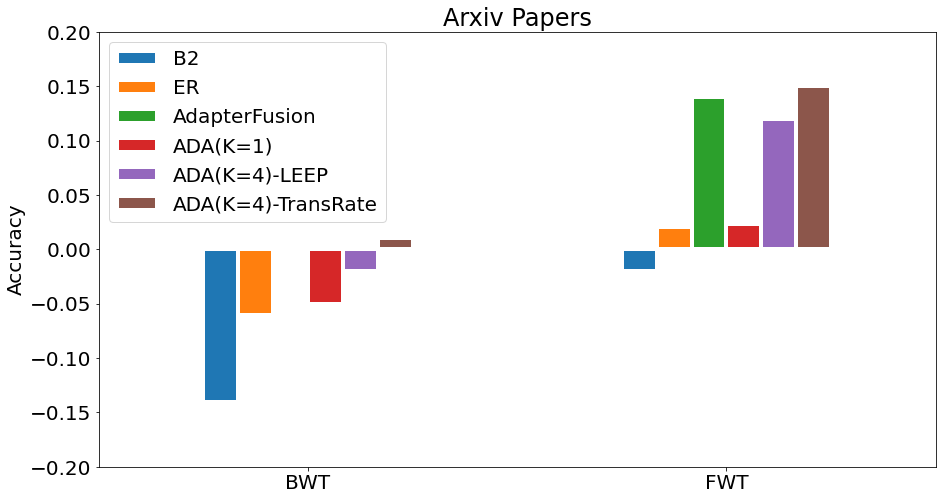}
\end{minipage}
\begin{minipage}{0.33\textwidth}
\centering
\includegraphics[width=\textwidth] {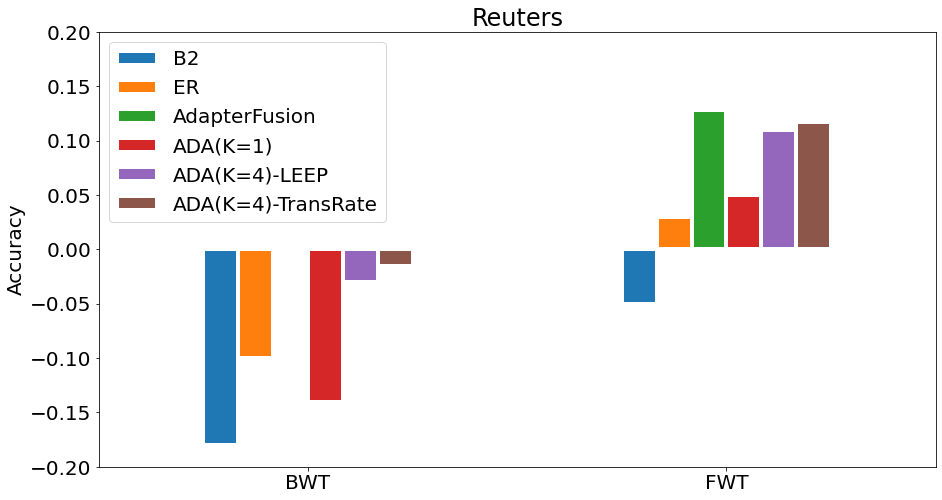}
\end{minipage}
\begin{minipage}{0.33\textwidth}
\centering
\includegraphics[width=\textwidth] {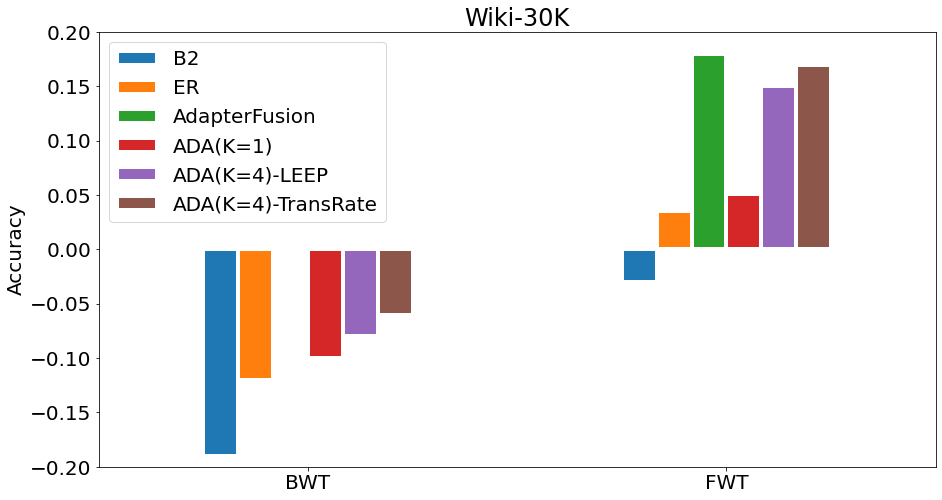}
\end{minipage}
\caption{Comparison between baselines and \al{} on Arxiv, Reuters and Wiki-30K.
On top, we report the number of tasks processed on the x-axis and we report the average accuracy measured 
on the test set of the tasks processed on the y-axis, shaded area shows standard deviation. On bottom, we report BWT and FWT.}
\label{fig:compAccuracies}
\end{figure}

\textbf{Adapter architectures.} 
We use pre-trained models from HuggingFace Transformers~\cite{wolf2020transformers} as our base feature extractors. 
We ran experiments with $\text{BERT}_{base}$, $\text{DistilBERT}_{base}$, $\text{RoBERTa}_{base}$ for text classification
and ViT-B and DeiT-B for image classification. We analyze the cases based on all these models, due to the space constraints,
we present $\text{BERT}_{base}$ in this section and the rest in Appendix~\ref{app:memory}. 
$\text{BERT}_{base}$ model uses 12 layers of Transformers block with a hidden size of 768 and number of self-attention heads
as 12 and has around 110 M (440 MB) trainable parameters.
For the Adapter implementation, we use Adapter-Hub~\cite{pfeiffer2020adapterhub}, but no Adapter implementation was available
for Vision Transformers. We define our architecture of Adapters for ViT and DeiT in Appendix~\ref{app:vis_adptr}.
An Adapter has a simple bottleneck architecture that contains fewer parameters than the attention and the
feed-forward layers. The Adapter size is the hyper-parameter that is tuned and it can be set to 
$\lbrace 12, 24, 48, 96, 192, 384\rbrace$ for $\text{BERT}_{base}$ model.
For all the methods, we use the same configuration for the Adapters, setting the size to $48$.
With this setting, an Adapter contains $\sim 1.8$ M parameters.
We also train a head model for each task, that has 768 parameters for $\text{BERT}_{base}$ 
(last hidden size of $\text{BERT}_{base}\times$output size, which equals to 1 for binary classification).
The tables in Appendix~\ref{app:memory} reports the number of parameters used for baselines and \al{} in our experiments.

\begin{figure}[ht!]
\begin{minipage}{0.33\textwidth}
\centering
\includegraphics[width=\textwidth] {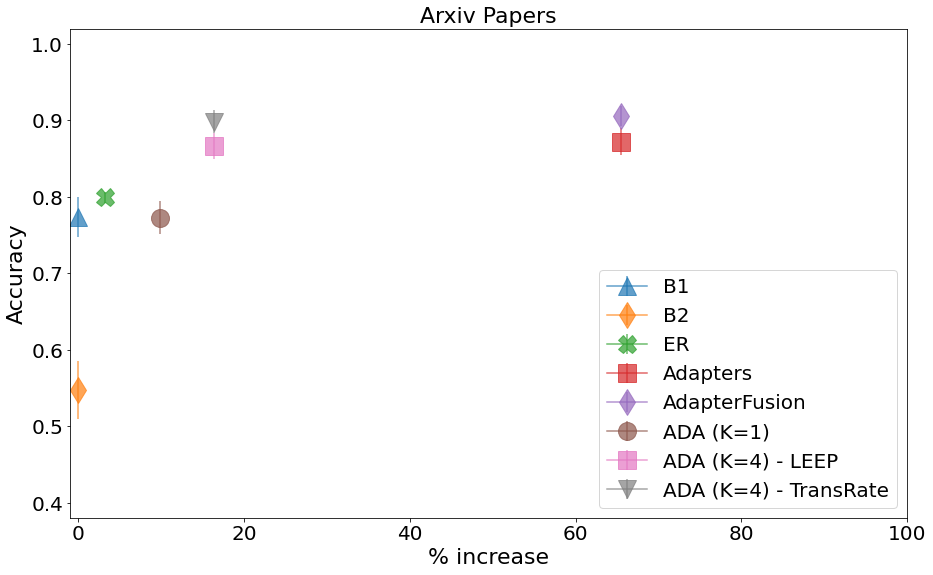}
\end{minipage}
\begin{minipage}{0.33\textwidth}
\centering
\includegraphics[width=\textwidth] {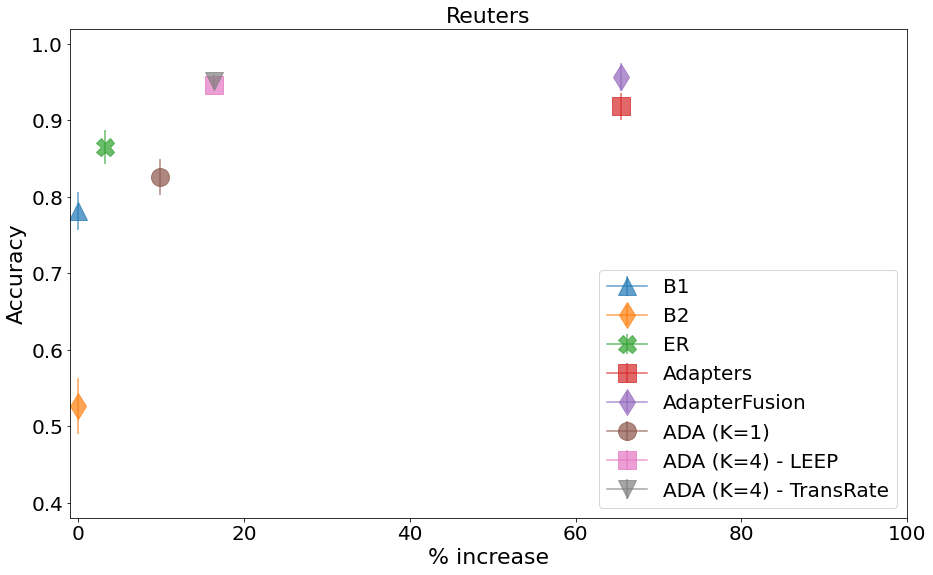}
\end{minipage}
\begin{minipage}{0.33\textwidth}
\centering
\includegraphics[width=\textwidth] {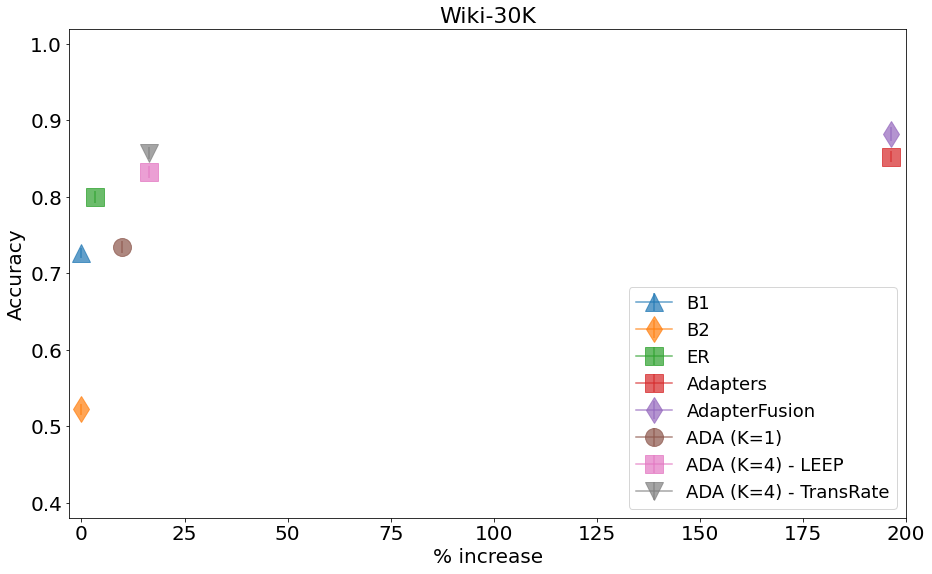}
\end{minipage}
\caption{Comparison of the $\%$ increase in the number of parameters of baseline methods and ADA on Arxiv, Reuters and Wiki-30K. 
The predictive performance reported on the y-axis is measured after processing all tasks.}
\label{fig:compMemoryPerc}
\end{figure}

\subsection{Text Classification}
\label{sec:text_classification}

\textbf{Predictive performance.}
Figure~\ref{fig:compAccuracies} shows the comparison of \al{} and the baseline methods. 
It can be clearly seen that freezing all pre-trained model parameters, and
fine-tuning only the head models (B1) led to an inferior performance compared to adapter-based approaches. 
The main reason is that the head models have small
amount of parameters to train and fine-tuning only the heads suffers from under-fitting. 
B2 performs good only for first 2-3 tasks, since we keep training the
complete model, it forgets the previously learned tasks very quickly. 
As mentioned above, Adapters and AdapterFusion add $\sim 1.8$ M parameters for each task and
train these parameters with new task data, and these parameters are fixed after training. 
So, they perform well on both new tasks and previous tasks. The results
on each dataset confirm this. 
Both ER and ADA K=1, perform closely with Adapters almost for half of the tasks. 
The similar behavior of ER and ADA K=1 demonstrates
that the distillation with soft labels works well and it is almost as good as training with the true labels. 
Later the performance declines for both methods, because
the capacity of the Adapter is exceeded. 
ADA LEEP and ADA TransRate results with K=4 Adapters show that selective consolidation of Adapters significantly improves
the performance. Their performance is on par with AdapterFusion while the number of model parameters is significantly lower. 
We present $\text{BERT}_{base}$ results in this section while the rest is reported in Appendix~\ref{a:db_roberta}.

We also compute FWT and BWT scores for these methods. We didn't present B1 and Adapters in the plots, since both FWT and BWT are zero for them.
BWT is zero for AdapterFusion, since the fusion parameter is computed with available Adapters, and the Adapters trained later is not used for
the previous tasks. ADA-LEEP and ADA-TransRate minimizes negative backward transfer, while showing a positive forward transfer for all datasets.

\textbf{Memory consumption.}
Figure~\ref{fig:compMemoryPerc} shows the increase in terms of percentage in the number of parameters used by each method and their predictive performance.
We see that on Wikipedia, $200\%$ of the base model parameters ($~220M$ additional parameters) are added. These results make clear that \al{} is significantly
more efficient in terms of memory usage. It can achieve predictive performance similar to the one of Adapters and AdapterFusion while requiring
significantly less model parameters. On Reuters and Arxiv, it can store the parameters of only 5 Adapters (K=4 Adapters in the pool, and one Adapter for new task),
against the 20 required by AdapterFusion (on Wikipedia it is 5 against 60).

\textbf{Inference time.}
When machine learning models are used to power customer-facing web sites, they are often required to provide predictions in a few 
milliseconds to keep the overall latency within requirements. Moreover, in this kind of application the model will be trained
once and make billions of predictions so a reasonable increase in the training time is irrelevant compared to a decrease in the
inference time. We report the inference time results of \al{} and other Adapter based methods in Appendix~\ref{a:inf_time}.
Results demonstrate that \al{} provides a sufficiently fast inference for most applications and still offers opportunities to speed
it up further, for example by employing smaller pre-trained Transformers (e.g. DistilBERT, see Appendix~\ref{app:memory}).

\textbf{Training time.} Distillation of Adapters brings an extra cost for ADA while learning fusion parameters brings an extra cost for AdapterFusion.
Computing transferability takes constant time which is negligible. Distillation costs training an additional Adapter (~1.6 \% of full fine-tuning time of BERT).
Figure~\ref{fig:wiki_train} in Appendix\ref{a:inf_time} reports the average training time comparison on Wiki-30K that is the largest difference with AdapterFusion
given larger number of tasks. We can clearly see that the difference is small (\al is 3.37\% more, ADA-TransRate is 5.6\% more) 
while the difference between the inference time is significant.

\subsection{Image Classification}
\label{sec:image_classification}

For image classification experiments, we add \emph{Elastic Weight Consolidation (EWC)}~\cite{kirkpatrick2017overcoming} as an additional baseline since it is 
widely used in CL literature for image classification. EWC is a regularization-based CL method that assumes that some weights of the trained neural 
network are more important for previously learned tasks than others. During training of the neural network on a new task, changes to the weights of the network
are made less likely the greater their importance.

\begin{figure*}[ht!]
\centering
\begin{minipage}{0.45\textwidth}
\includegraphics[width=\textwidth]{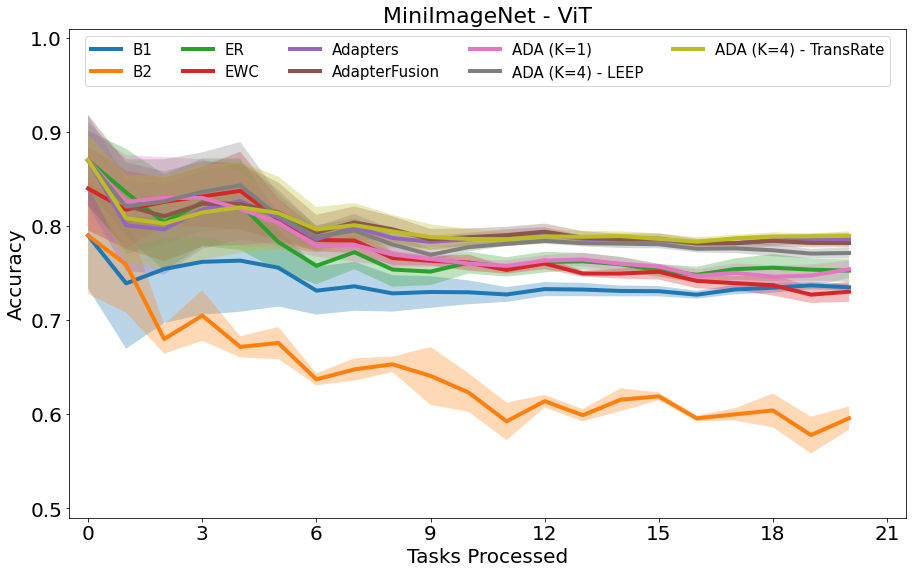}
\end{minipage}
\hspace*{5mm}
\begin{minipage}{0.45\textwidth}
\includegraphics[width=\textwidth]{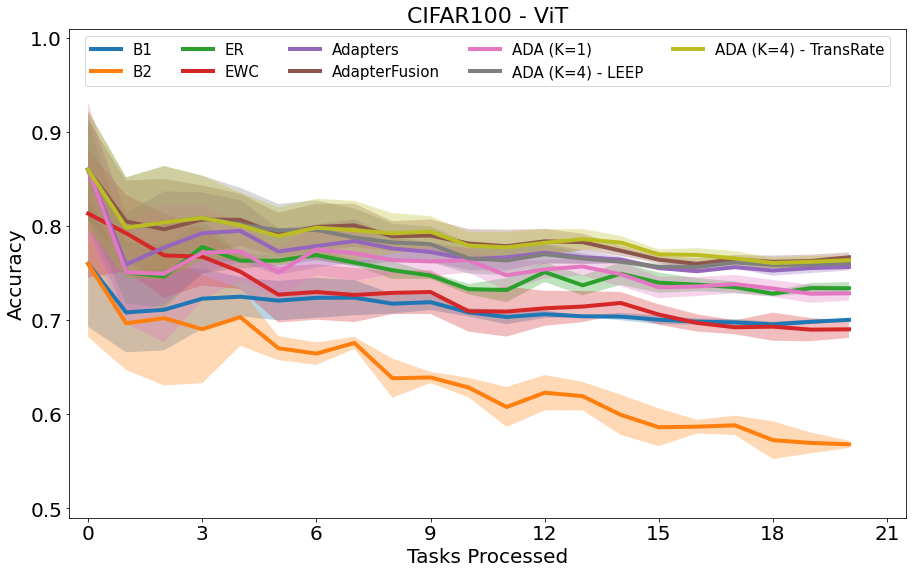}
\end{minipage}
\begin{minipage}{0.45\textwidth}
\includegraphics[width=\textwidth]{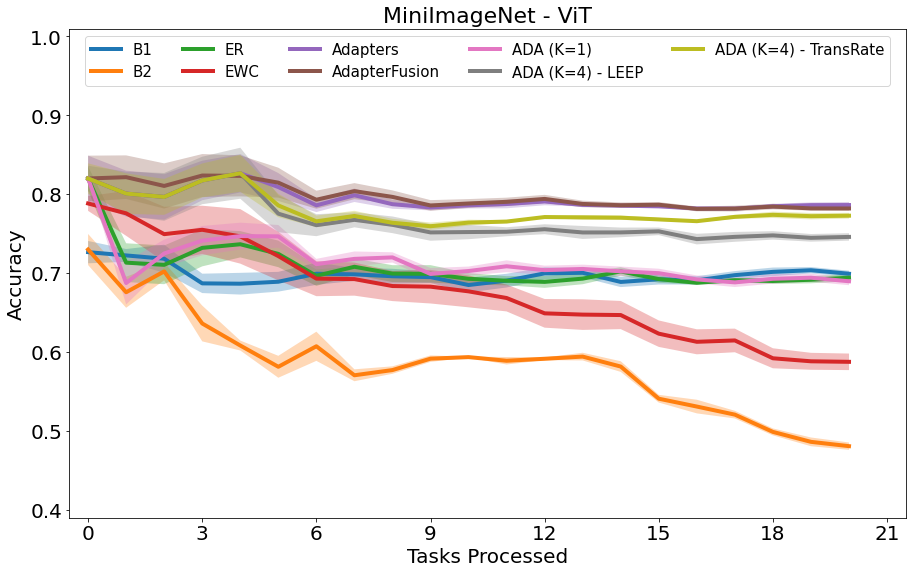}
\end{minipage}
\hspace*{5mm}
\begin{minipage}{0.45\textwidth}
\includegraphics[width=\textwidth]{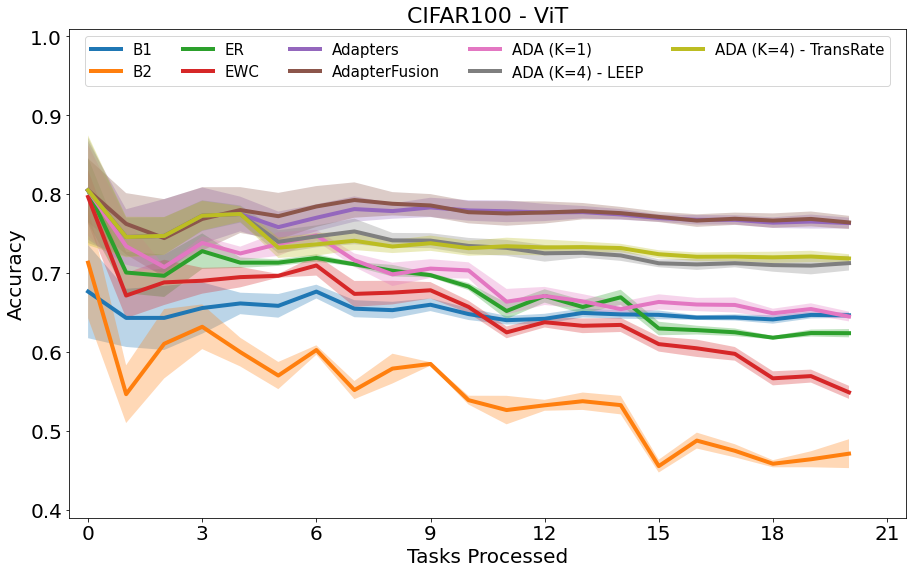}
\end{minipage}
\caption{Comparison between baselines and ADA with ViT model on MiniImageNet and CIFAR100.
Top figures shows the binary, and bottom figures shows the multi-class classification results.}
\label{fig:compAccuraciesViT}
\end{figure*}
Figure~\ref{fig:compAccuraciesViT} shows the comparison of ADA and the baseline methods.
The results show the same behaviour with text classification. B1 leaded to an inferior performance compared to other approaches. 
B2 performs well only for initial tasks and it forgets the previously learned tasks very quickly. 
Results confirm there is no forgetting for Adapters and AdapterFusion. Although the careful tuning of regularization coefficient,
EWC cannot handle CF, especially for multi-class classification problem.
ADA with K=1 shows that distillation alone doesn't prevent forgetting. In almost all cases, ER performs on-par with ADA K=1, 
providing evidence that a small amount of memory can actually improve performance compared to fine-tuning or regularization,
but the improvement is limited and does not last as the number of tasks increases.

ADA-LEEP and ADA-TransRate results with K=4 Adapters show that selective consolidation of Adapters significantly improves
the performance. For binary classification, their performance are on par with AdapterFusion while the number of model
parameters is significantly lower. For multi-class, their performance slightly declines after a certain number of tasks.
This is discussed in next section and the main reason is that the capacity of the Adapter is exceeded.
To validate the interoperability of ADA to different models, we run the same experiments on DeiT model and present the results
in Appendix~\ref{a:deit_exp} due to space constraints.

\subsection{Ablation studies}
\label{sec:ablation}

\textbf{Comparison with larger distilled models.}
In Section~\ref{sec:experiments} we compared \al{} with the special case of \al{} with K=1 to evaluate the 
improvement provided by our approach over a distillation-only solutions.
We would like to provide additional observations of the superior performance of \al{} by
comparing its performance with the one of a distilled Adapter using more parameters.
Specifically, we run an experiment where we compare \al{} with K=4 and \al{} with K=1 as displayed before
but in this case the ``size'' of the Adapter, which is 48 for Size$\times$1, is multiplied by 4 for Size$\times$4 Adapter
to have a comparison where the different methods use the same number of model parameters.
Since K=1 is a special case where a single Adapter is kept in the pool, the transferability metric is irrelevant
and we can see \al{} with K=1 as a method purely based on distillation like DMC~\cite{zhang2020class}.

\begin{figure}[ht!]
\centering
\begin{minipage}{0.45\textwidth}
\includegraphics[width=\textwidth] {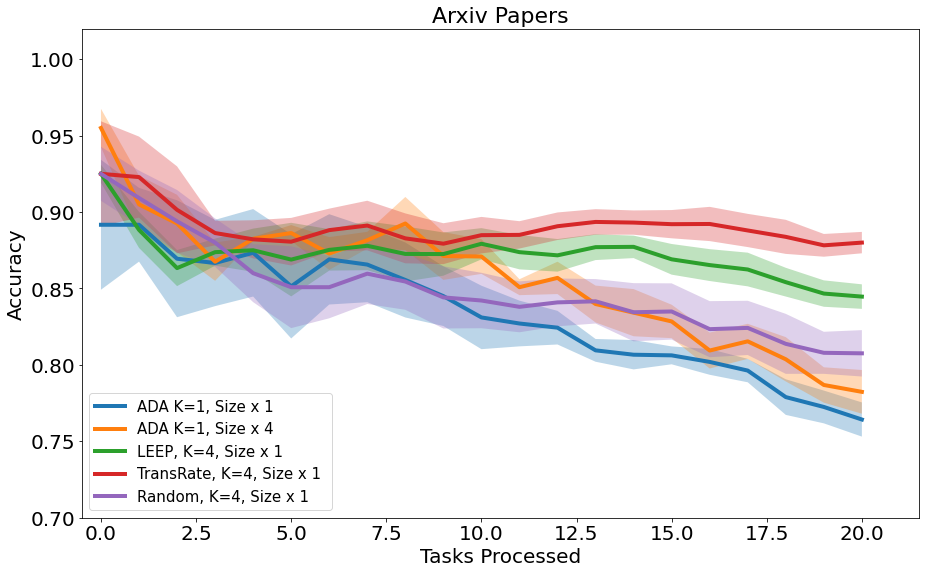}
\end{minipage}
\hspace*{5mm}
\begin{minipage}{0.45\textwidth}
\includegraphics[width=\textwidth] {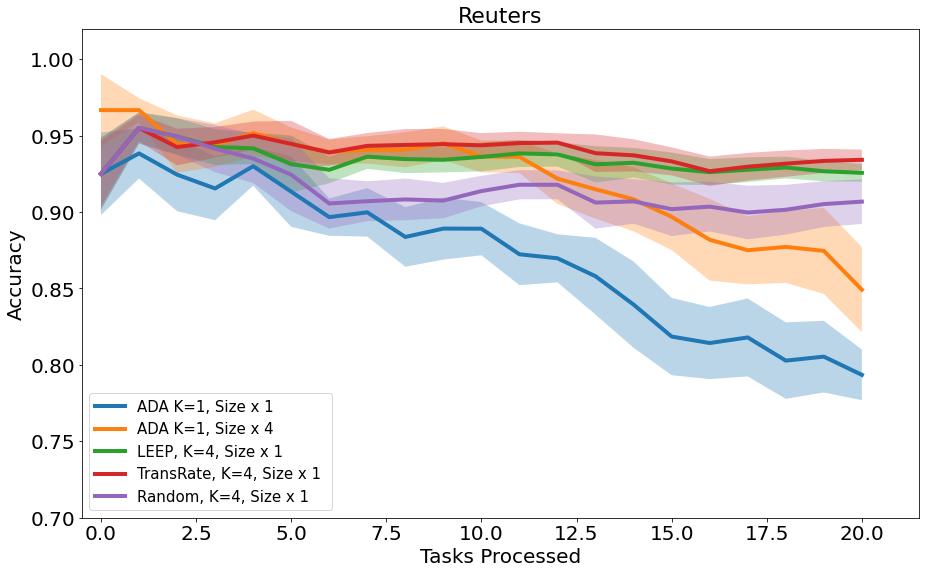}
\end{minipage}
\caption{Impact of \emph{LEEP} and \emph{TransRate} when the total number of Adapter parameters is same on Arxiv and Reuters.}
\label{fig:ADA_size_comp}
\end{figure}

The results reported in Figure~\ref{fig:ADA_size_comp} show that \al{} can make a better use of the model parameters
compared to a distillation-only method and that the intelligent selection of which Adapters to distill together
makes once again a big difference. It is also interesting to observe that the usage of additional model parameters
brings a clear advantage but the mixed comparison between the \al{} K=4 with random Adapter selection and \al{} K=1 with 
four times larger Adapters leaves some questions open regarding how far distillation can get in this setting.
Another finding is that \emph{TransRate} outperforms \emph{LEEP} in most cases. It is also demonstrated in the original
paper~\cite{huang2021frustratingly} that \emph{TransRate} has a strong correlation to the transfer learning performance
and it outperforms \emph{LEEP} and other metrics employed.

\textbf{Impact of the Adapters pool size.}
In our experiments we used a fixed number of Adapters in the pool size, but more Adapters can be added 
to \al{}'s pool as more tasks are processed. This may actually be the preferred usage in some applications.
We already know that having an Adapter per task provides good performance and using multiple of them 
at the same time like in AdapterFusion provides a benefit, but we would like to verify the sensitivity to this parameter.
\begin{figure}[ht!]
\centering
\begin{minipage}{0.45\textwidth}
\includegraphics[width=\textwidth] {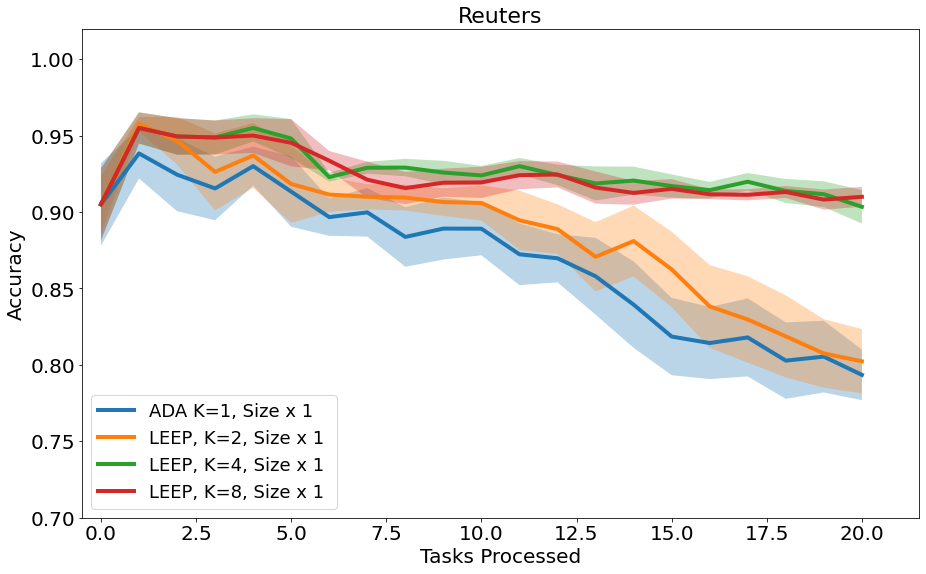}
\end{minipage}
\hspace*{5mm}
\begin{minipage}{0.45\textwidth}
\includegraphics[width=\textwidth] {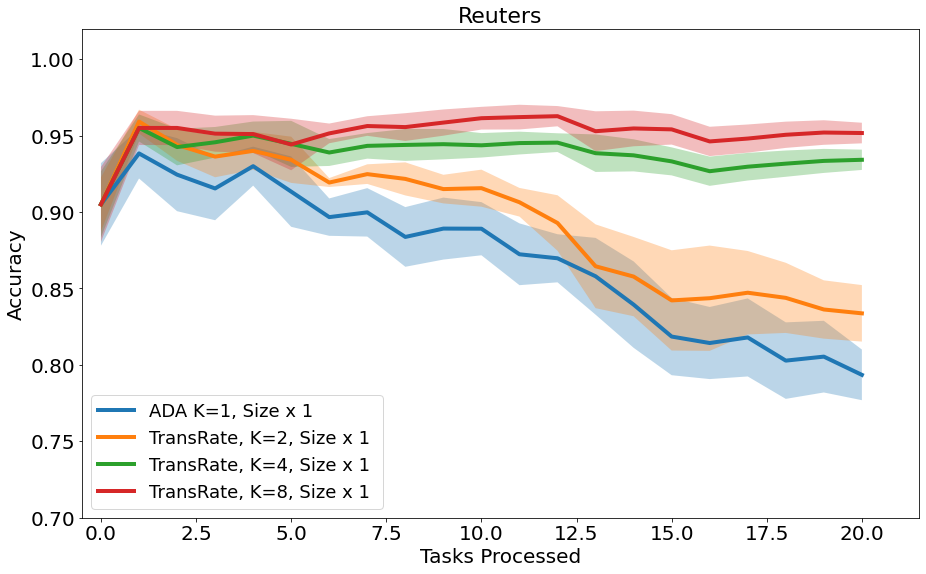}
\end{minipage}
\caption{\emph{LEEP} and \emph{TransRate} performances when $K=\lbrace 1,2,4,8 \rbrace$ on Reuters.}
\label{fig:ADA_Kcomp}
\end{figure}
The results reported in Figure~\ref{fig:ADA_Kcomp} show a rapidly decreasing added value when the number of Adapters grows,
a behavior which aligns well with our practical requirements of keeping the number of model parameters under control when
the number of tasks increases. See additional experiments in Appendix~\ref{a:addpoolsize}.

\begin{figure}[ht!]
\centering
\begin{minipage}{0.45\textwidth}
\includegraphics[width=\textwidth] {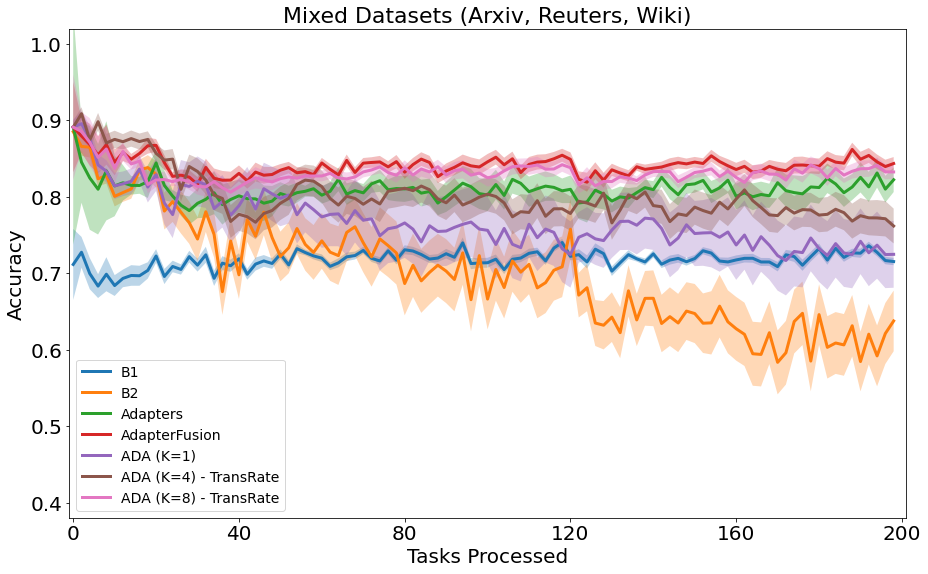}
\end{minipage}
\hspace*{5mm}
\begin{minipage}{0.45\textwidth}
\includegraphics[width=\textwidth] {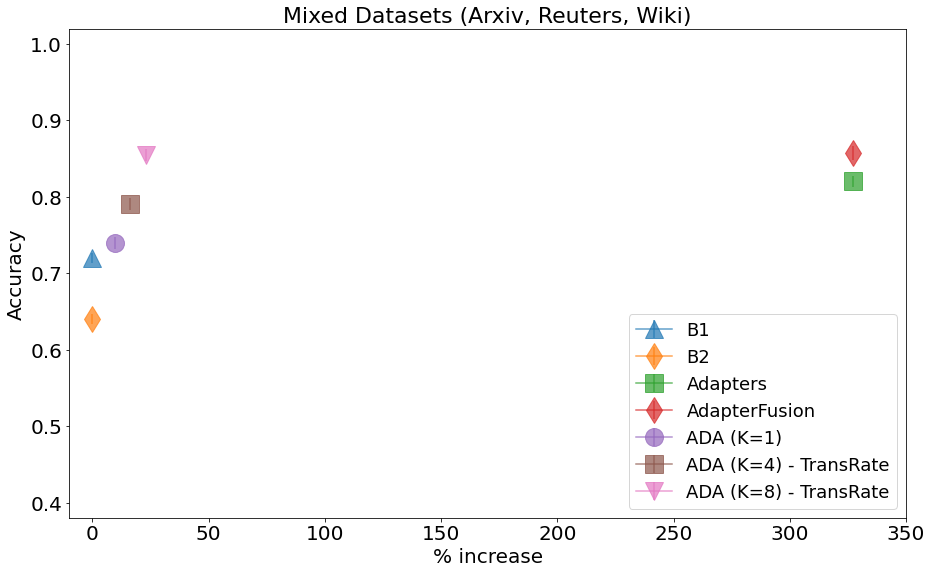}
\end{minipage}
\caption{(Left) method performance comparisons (Right) comparison of the $\%$ increase in the number of parameters of baseline methods and ADA on mix of datasets.
\emph{LEEP} shows a similar performance with \emph{TransRate}, for the sake of clarity, we didn't add it to the figures.}
\label{fig:ADA_mixed}
\end{figure}

\textbf{Mixed Data Experiments.}
We run experiments in a setting where we sample 200 tasks from Arxiv, Reuters and Wikipedia (50/50/100) respectively (the order of the tasks are created randomly).
We fixed the number of training and test samples per task to 100.
Figure~\ref{fig:ADA_mixed} shows that we observe a little saturation only after the 150th task when $K$=4 and no saturation when $K$=8. Besides, \al{} with \emph{TransRate}
comparable performance with Adapters and AdapterFusion even in a complicated setting. Figure~\ref{fig:ADA_mixed} also shows the increase in terms of percentage in the number
of parameters used by each method and their predictive performance. We see that $\sim 330\%$ of the base model parameters are added for Adapters and AdapterFusion.
These results make clear that \al{} is significantly more efficient in terms of memory usage while keeping the comparable performance.

\section{Conclusion}
\label{sec:conclusion}
\vspace*{-2mm}
In this paper we presented \al{}, a method that allows neural text and image
classifiers to learn new classes based on pre-trained Transformers while maintaining strict control of the memory usage and reaching state-of-the-art predictive performance.
The method has shown to be effective in different domains and allows users to leverage publicly available 
pre-trained Transformers for continual classification tasks.
We evaluated \al{} on different classification tasks and demonstrated that the predictive performance is
competitive with state-of-the-art methods which use up to an order of magnitude parameters. 
\al{} also displayed lower
latency at inference time and improved data efficiency for some specific settings (see Appendix~\ref{a:dataeff}).
Moreover, we empirically demonstrated that Adapters can give good results when used in combination with vision Transformers on CV tasks.

Transformers are very popular, but they are not the only models being widely used in practice.
We consider this the main weakness of our approach and we would like to further expand our activity 
to perform CL on other widely used pre-trained models such as ResNet.
Addressing multi-modal classification using text and images together will be the other focus of our future research.

\medskip
\small
\bibliographystyle{plain}
\bibliography{neurips_2022}

\newpage
\section*{Checklist}

The checklist follows the references.  Please
read the checklist guidelines carefully for information on how to answer these
questions.  For each question, change the default \answerTODO{} to \answerYes{},
\answerNo{}, or \answerNA{}.  You are strongly encouraged to include a {\bf
justification to your answer}, either by referencing the appropriate section of
your paper or providing a brief inline description.  For example:
\begin{itemize}
  \item Did you include the license to the code and datasets? \answerYes{See Section~\ref{gen_inst}.}
  \item Did you include the license to the code and datasets? \answerNo{The code and the data are proprietary.}
  \item Did you include the license to the code and datasets? \answerNA{}
\end{itemize}
Please do not modify the questions and only use the provided macros for your
answers.  Note that the Checklist section does not count towards the page
limit.  In your paper, please delete this instructions block and only keep the
Checklist section heading above along with the questions/answers below.

\begin{enumerate}

\item For all authors...
\begin{enumerate}
  \item Do the main claims made in the abstract and introduction accurately reflect the paper's contributions and scope?
    \answerYes{}
  \item Did you describe the limitations of your work?
    \answerYes{Our work is tied to the usage of Transformers, we remind it in the conclusions.}
  \item Did you discuss any potential negative societal impacts of your work?
    \answerNA{}
  \item Have you read the ethics review guidelines and ensured that your paper conforms to them?
    \answerYes{}
\end{enumerate}

\item If you are including theoretical results...
\begin{enumerate}
  \item Did you state the full set of assumptions of all theoretical results?
    \answerNA{We don't have theoretical results.}
        \item Did you include complete proofs of all theoretical results?
    \answerNA{We don't have theorems.}
\end{enumerate}

\item If you ran experiments...
\begin{enumerate}
  \item Did you include the code, data, and instructions needed to reproduce the main experimental results (either in the supplemental material or as a URL)?
    \answerTODO{We will provide the code if the paper is accepted.}
  \item Did you specify all the training details (e.g., data splits, hyperparameters, how they were chosen)?
    \answerYes{See Section~\ref{sec:experiments} and Appendix A.2.}
        \item Did you report error bars (e.g., with respect to the random seed after running experiments multiple times)?
    \answerYes{Yes, all figures show the standard deviation (shaded regions).}
        \item Did you include the total amount of compute and the type of resources used (e.g., type of GPUs, internal cluster, or cloud provider)?
    \answerYes{See Appendix A.2.}
\end{enumerate}

\item If you are using existing assets (e.g., code, data, models) or curating/releasing new assets...
\begin{enumerate}
  \item If your work uses existing assets, did you cite the creators?
    \answerYes{We used AdapterHub, that is cited in experiments and appendix sections.}
  \item Did you mention the license of the assets?
    \answerYes{Yes, in Appendix A.3.}
  \item Did you include any new assets either in the supplemental material or as a URL?
    \answerNA{}
  \item Did you discuss whether and how consent was obtained from people whose data you're using/curating?
    \answerNA{}
  \item Did you discuss whether the data you are using/curating contains personally identifiable information or offensive content?
    \answerNA{}
\end{enumerate}

\item If you used crowdsourcing or conducted research with human subjects...
\begin{enumerate}
  \item Did you include the full text of instructions given to participants and screenshots, if applicable?
    \answerNA{}
  \item Did you describe any potential participant risks, with links to Institutional Review Board (IRB) approvals, if applicable?
    \answerNA{}
  \item Did you include the estimated hourly wage paid to participants and the total amount spent on participant compensation?
    \answerNA{}
\end{enumerate}

\end{enumerate}

\newpage
\appendix

\section{Memory Efficient Continual Learning with Transformers: Appendix}
\label{app:appendix}

\subsection{Related work on CL approaches, distillation and transferability}
\label{app:related}

\paragraph{Continual Learning (CL).}
Existing methods for CL can be roughly categorized as follows:
(1) Replay-based methods~\citep{lopez2017gradient,rolnick2018experience,d2019episodic,chaudhry2019continual,wang2020efficient} 
retain some training data of old tasks
and use them in learning a new task to circumvent the issue of catastrophic forgetting (CF); 
(2) Regularization-based methods~\citep{kirkpatrick2017overcoming,aljundi2018selfless,huang2021continual} 
add a regularization term to the loss to consolidate previous 
knowledge when learning a new task; 
(3) Gradient-based methods~\citep{zeng2019continual,aljundi2019gradient} ensure the gradient updates occur only 
in the orthogonal direction to the input of old tasks and thus 
will not affect old tasks. Recently, some studies use pre-trained models for class incremental learning~\citep{ke2021adapting,hu2021continual}; 
(4) Parameter isolation-based methods~\citep{ke2020continual,wortsman2020supermasks} 
allocate model parameters dedicated to different tasks and mask them out when learning a new task; 
(5) Meta-learning-based methods, which directly optimize the knowledge transfer among tasks~\citep{riemer2018learning,obamuyide2019meta} 
or learn robust data representations~\citep{javed2019meta,wang2020efficient}.

\paragraph{Distillation for CL.} Knowledge distillation refers to the process of transferring the knowledge from a large bulky model or a set of models to a 
single smaller model that can be practically deployed under real-world constraints. Essentially, it is a form of model compression that was first proposed
by~\cite{bucilua2006model} and used by~\cite{hinton2015distilling} to preserve the output of a complex ensemble of networks when adopting a simpler network
for more efficient deployment. 
The idea is adopted in CL and incremental learning domain to maintain the responses of the network unchanged on the old tasks whilst updating it with new training samples in different 
ways~\citep{shmelkov2017incremental,castro2018end,li2017learning,zhou2019m2kd}. \cite{shmelkov2017incremental} propose an end-to-end learning framework where the representation and 
the classifier are learned jointly without storing any of the original training samples. \cite{li2017learning} distill previous knowledge directly from the last trained model. 
\cite{zhou2019m2kd} propose to use the current model to distill knowledge from all previous model snapshots, of which a pruned version is saved. \cite{schwarz2018progress} use distillation to consolidate the network after each task has been learned and \cite{buzzega2020dark} leverage knowledge distillation for retaining past experience.

We inspired from the idea proposed by Zhang et al. \cite{zhang2020class} where two individual image classification models trained on image data of two distinct set of classes
(old classes and new classes) are consolidated into one single model that can classify all classes. The training objective for consolidation is defined as:
\begin{align}
\min_\Theta \frac{1}{\mathcal{U}} \sum_{x_i \in \mathcal{U}} L_{dd}(\bm{y}_i, \bm{\hat{y}}_i)
\end{align}
where $\mathcal{U}$ denotes the unlabeled auxiliary training data and the double distillation loss $L_{dd}$ is defined as:
\begin{align}
L_{dd}(\bm{y}_i, \bm{\hat{y}}_i) = \frac{1}{t} \sum_{j=1}^t(y^j - \hat{y}^j)^2
\end{align}
in which $y^j$ is the logit produced by the consolidated model for the $j$-th class.
In our work, we adopt the idea of model consolidation and use it for
incremental text classification. In our setting, we leverage the pre-trained model, keep it fixed, and only use Adapters to transfer knowledge from old tasks to the new tasks
and train one Adapter that can perform well on all classification tasks.
Our main goal is to use the advantage of knowledge transfer between tasks with distillation. So we also use transferability estimation methods to select the Adapters that needs
to be distilled. By enhancing the power of distillation, we achieve the same performance with state-of-the-art methods while keeping the number of model parameters much smaller.

\paragraph{Task Transferability.} Automatically selecting intermediate tasks that yield transfer gains is critical when considering the increasing availability
of tasks and models. There are a number of works that explores task transferability in NLP~\citep{phang2018sentence,liu2019linguistic,vu2020exploring,puigcerver2020scalable,pruksachatkun2020intermediate}. Poth et al.~\cite{poth2021pre} present
a large-scale study on Adapter-based sequential fine-tuning. Given multiple source and target task pairs ($s$,$t$), they first train an Adapter on $s$,
then fine-tune the trained Adapter on $t$ and show the relative transfer gains across the different combinations. They use different methods for intermediate task selection,
and LEEP~\citep{nguyen2020leep} is one of the methods that they used in this work to measure transferability and it is consolidated in NLP domain.
TransRate~\citep{huang2021frustratingly} is a very recent work and it is used with image classification tasks in the original work. To the best of our knowledge, we use TransRate
for the first time in NLP domain. Our work is quite different from what is proposed in the literature. We focus on selecting the best representation from a pool of representations 
(trained Adapters) for model consolidation, without the necessity of computationally expensive additional approach. We use \emph{proxy estimators}, LEEP and TransRate, that 
evaluate the transferability of pre-trained models towards a target task without explicit training on all potential candidates.

\subsection{ADA algorithm}
\label{app:algo}

\begin{figure}[ht!]
\centering
\includegraphics[scale = 0.15] {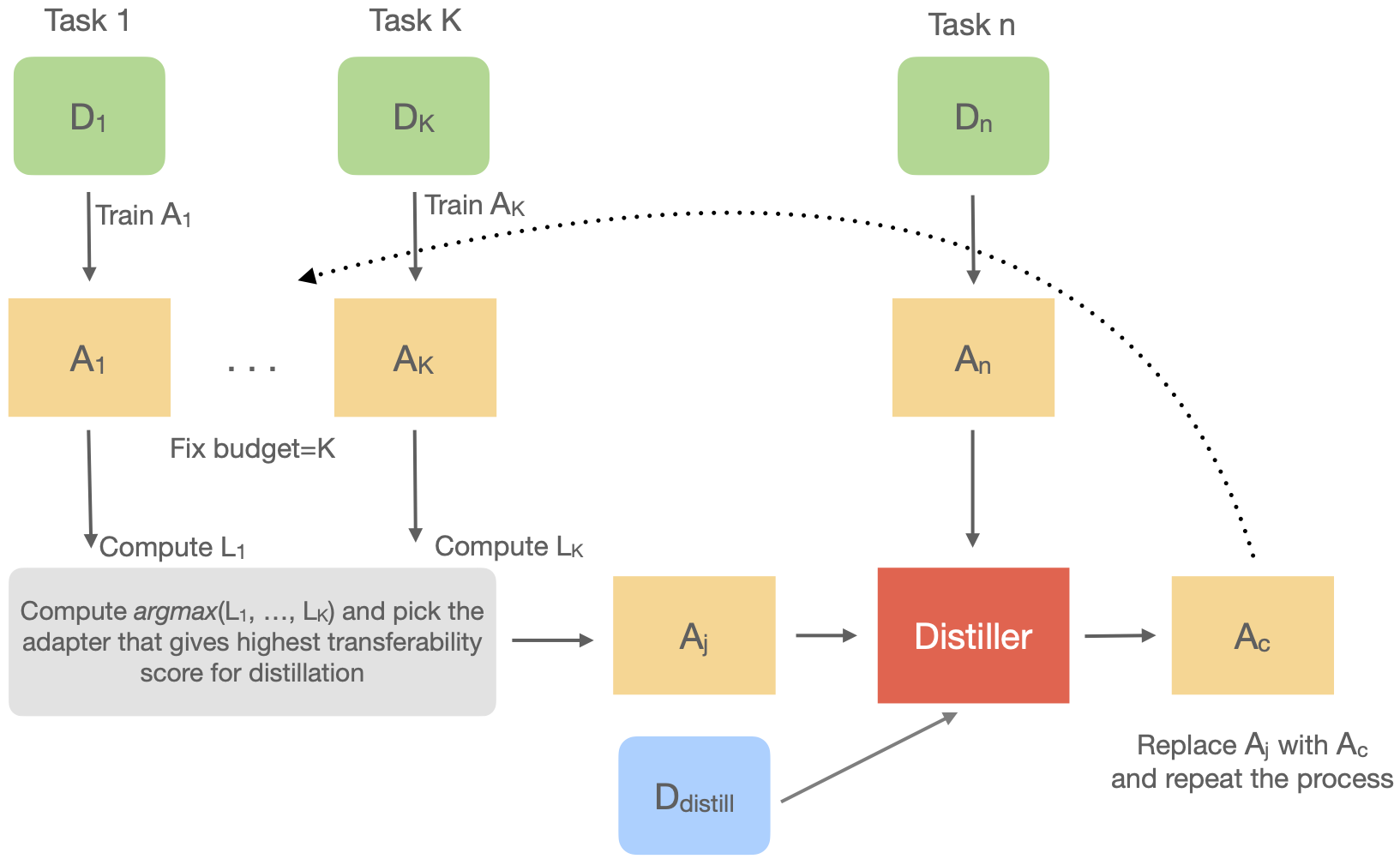}
\caption{ADA workflow.}
\label{fig:ADAalg}
\end{figure}

The visual representation of \al{} is shown in Figure~\ref{fig:ADAalg}.

\subsection{Datasets and Experimental Setup}
\label{app:datasets}

\textbf{Datasets.} Arxiv papers dataset contains the abstract and the corresponding subjects of 55,840 papers in the computer science field from Arxiv.org. 
There are 54 subjects in total and each paper can cover multiple subjects. In our work each of these subjects will represent a different task for the classifier
where the target is to predict corresponding subjects of an academic paper according to the content of the abstract.
Reuters consists of over 800,000 manually categorized newswire stories made available by Reuters Ltd for research purposes. 
Multiple topics can be assigned to each newswire story and there are 103 topics in total. For Wiki-30K, a set of tags for the English Wikipedia was gathered. 
Starting with a set of more than 2 million articles from the English Wikipedia on April 2009, the tag information for each of these articles was retrieved from
the social bookmarking site Delicious. Only the articles annotated by at least 10 users in Delicious were preserved.
As a result, a dataset with 20,764 tagged Wikipedia articles was generated. There are 29,947 labels in this dataset.
Both CIFAR100~\cite{krizhevsky2009learning} and MiniImageNet~\cite{russakovsky2015imagenet} consist of 60000 colour images in 100 classes, with 600 images per class.

\textbf{Setup.} We use \emph{Adam} as optimizer with the batch size of 8. For learning rate, we select best from 
$\lbrace 0.00005, 0.0001, 0.0005, 0.001\rbrace$ after observing the results on the first five tasks.\\
We tune the regularization coefficient of EWC by grid search in $\lbrace 0,1,10,100,1000\rbrace$.

As computation environment, we used Amazon G4dn instances that provide up to 8 NVIDIA T4 GPUs, 96 vCPUs, 100 Gbps networking, and 1.8 TB local 
NVMe-based SSD storage and are also available as bare metal instances.

\subsection{Adapter Architecture}
\label{app:vis_adptr}

\textbf{Architecture.} Figure below shows the Adapter architecture and it's integration with transformer. In~\cite{houlsby2019parameter},
they add the adapter module twice to each Transformer layer: after the projection following multi-headed attention and after the two feed-forward layers.
To limit the number of parameters, a bottleneck architecture is proposed. The adapters first project the original $d$-dimensional features into a smaller dimension,
$m$, apply a non-linearity, then project back to $d$ dimensions.
The total number of parameters added per layer, including biases, is $2md + d + m$. By setting $m \ll d$, the number of parameters added per task is limited.

\textbf{Vision Adapters.} One other contribution of this work is using Adapters approach with vision Transformers for the first time on sequential image
classification tasks, validating that Adapters work with vision Transformers and show that \al{} can achieve predictive performance
on-par with AdapterFusion. We implement vision Transformer Adapters in 
AdapterHub~\cite{pfeiffer2020adapterhub}\footnote{\url{https://github.com/Adapter-Hub/adapter-transformers}} (that has 
Apache License, Version 2.0). 
As in AdapterBERT~\cite{houlsby2019parameter}, we insert a 2-layer fully-connected network in each Transformer layer of
ViT~\cite{dosovitskiy2020image} and DeiT~\cite{touvron2021training} is built upon the ViT architecture, so an Adapter is added in the same way.
\begin{figure}[ht!]
\begin{minipage}{0.33\textwidth}
\centering
\includegraphics[scale = 0.15] {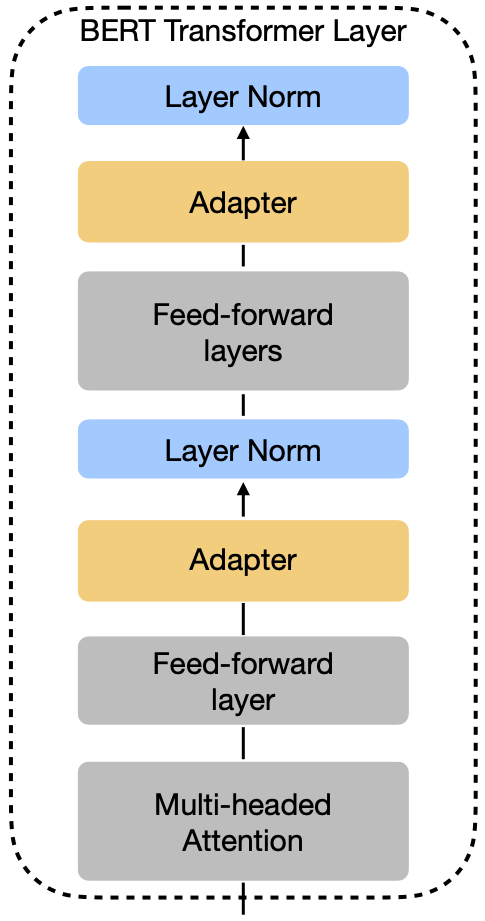}
\end{minipage}
\begin{minipage}{0.33\textwidth}
\centering
\includegraphics[scale = 0.15] {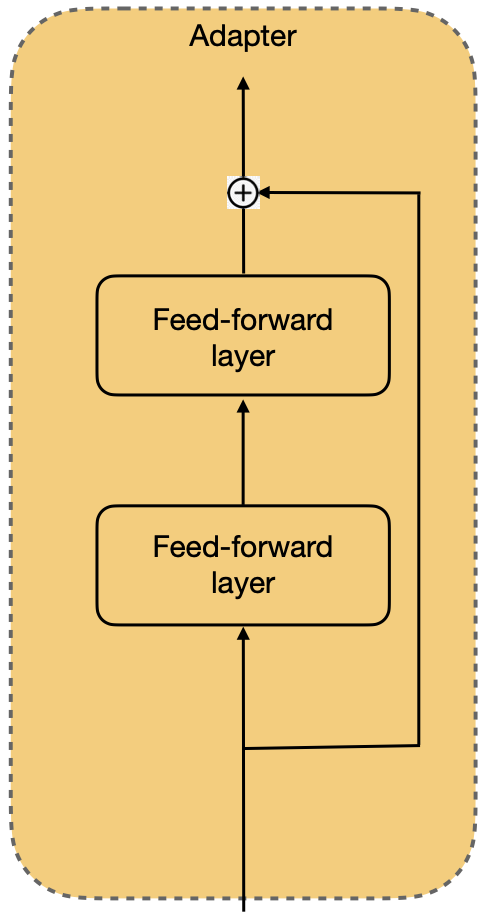}
\end{minipage}
\begin{minipage}{0.33\textwidth}
\centering
\includegraphics[scale = 0.15] {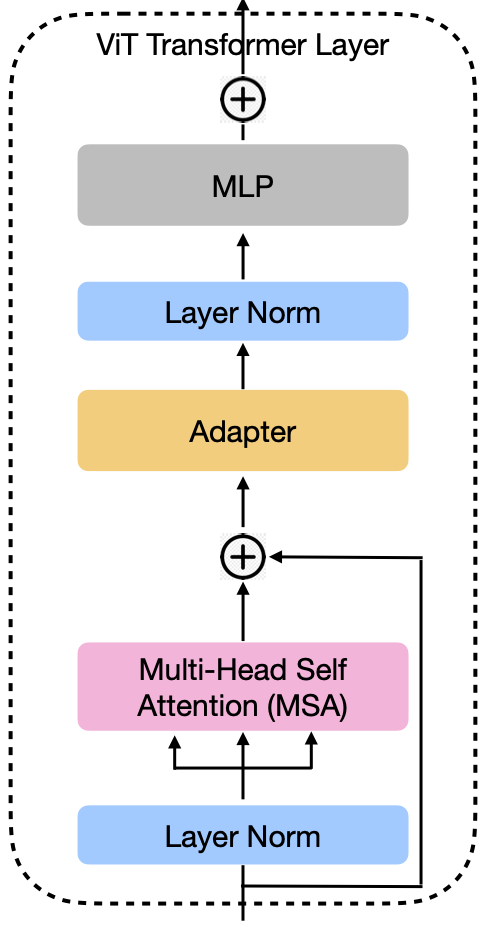}
\end{minipage}
\caption{Left shows AdapterBERT~\cite{houlsby2019parameter} in a BERT transformer layer, and middle shows the Adapter architecture.
Right shows our Adapter implementation in a ViT~\cite{dosovitskiy2020image} transformer layer. As in AdapterBERT,
we added an Adapter before layer norm and feed-forward layers (MLP).}
\label{fig:AdapterArch}
\end{figure}

\subsection{Trainable parameters for different models}
\label{app:memory}

The tables below reports the number of parameters used for baselines and \al{} in our experiments. We reported all the cases for different models: 
$\text{BERT}_{base}$, $\text{RoBERTa}_{base}$ and $\text{DistilBERT}_{base}$.
We don't add the head size to the table, since it's very small and same for all the methods. 

\begin{table}[H]
\caption{The number of all parameters and those used for training and inference as well as the model size of methods for $\text{BERT}_{base}$.
$K$ is the number of Adapters in the pool, and $F$ is the number of fused Adapters (it is between 2 and number of tasks). For the \emph{Adapters} $F=1$.}
\begin{minipage}[b]{1.0\hsize}\centering
\scalebox{0.80}{
\begin{tabular}{l | l | l | l | l }
           & \multicolumn{4}{c }{Fine-Tuning} \\  \hline
           & Trainable  & Inference  & Total    &  Total (Size)   \\ \hline \hline
Task = $\lbrace 1,10,30,60\rbrace$   & 110 M    &  110 M   &  110 M   &  440 MB   
\end{tabular}
}
\end{minipage}
\\
\begin{minipage}[b]{1.0\hsize}\centering
\vspace*{3mm}
\scalebox{0.82}{
\begin{tabular}{l | l | l | l | l }
           & \multicolumn{4}{c }{Adapters \& AdapterFusion}                       \\  \hline
           & Trainable  & Inference   & Total          & Total (Size)             \\ \hline \hline
Task = 1 \hspace*{2mm}  & 1.8 M       &  111.8 M  &  111.8 M  & 447.2 MB            \\ \hline
Task = 10  & 1.8 M  &  110 + (F$\times$1.8) M  &  128 M   & 512 MB                   \\ \hline
Task = 30  & 1.8 M  &  110 + (F$\times$1.8) M  &  164 M   & 656 MB                   \\ \hline
Task = 60  & 1.8 M  &  110 + (F$\times$1.8) M  &  218 M   & 872 MB
\end{tabular}
}
\end{minipage}
\\
\begin{minipage}[b]{1.0\hsize}\centering
\vspace*{3mm}
\scalebox{0.80}{
\begin{tabular}{l | l | l | l | l }
           & \multicolumn{4}{c }{ADA}                                       \\  \hline
           & Trainable  & Inference   & Total    & Total (Size)             \\ \hline \hline
Task = 1  & 1.8 M       &  111.8 M  &  111.8 M         & 447.2 MB   \\ \hline
Task = $\lbrace 10,30,60\rbrace$   &  2$\times$1.8 M  &  111.8 M   &  110 + (K+1)$\times$1.8 M  &  440 + (K+1)$\times$7.2 MB
\end{tabular}
}
\end{minipage}
\label{app:numparam}
\end{table}

\begin{table}[H]
\caption{The number of all parameters and those used for training and inference as well as the model size of methods for $\text{RoBERTa}_{base}$.}
\begin{minipage}[b]{1.0\hsize}\centering
\scalebox{0.80}{
\begin{tabular}{l | l | l | l | l }
           & \multicolumn{4}{c }{Fine-Tuning} \\  \hline
           & Trainable  & Inference  & Total    &  Total (Size)   \\ \hline \hline
Task = $\lbrace 1,10,30,60\rbrace$   & 125 M    &  125 M   &  125 M   &  500 MB   
\end{tabular}
}
\end{minipage}
\\
\begin{minipage}[b]{1.0\hsize}\centering
\vspace*{3mm}
\scalebox{0.82}{
\begin{tabular}{l | l | l | l | l }
           & \multicolumn{4}{c }{Adapters \& AdapterFusion}                       \\  \hline
           & Trainable  & Inference   & Total       & Total (Size)             \\ \hline \hline
Task = 1 \hspace*{2mm}  & 1.8 M       &  126.8 M    &  126.8 M   &   507.2 MB            \\ \hline
Task = 10  & 1.8 M  &  125 + (F$\times$1.8) M   &  143 M   & 584 MB                   \\ \hline
Task = 30  & 1.8 M  &  125 + (F$\times$1.8) M   &  179 M   & 716 MB                   \\ \hline
Task = 60  & 1.8 M  &  125 + (F$\times$1.8) M   &  233 M   & 932 MB
\end{tabular}
}
\end{minipage}
\\
\begin{minipage}[b]{1.0\hsize}\centering
\vspace*{3mm}
\scalebox{0.80}{
\begin{tabular}{l | l | l | l | l }
           & \multicolumn{4}{c }{ADA}                                       \\  \hline
           & Trainable  & Inference   & Total     & Total (Size)             \\ \hline \hline
Task = 1  & 1.8 M       &  126.8 M   &  126.8 M   & 507.2 MB   \\ \hline
Task = $\lbrace 10,30,60\rbrace$   &  2$\times$1.8 M  &  126.8 M   &  125 + (K+1)$\times$1.8 M  &  500 + (K+1)$\times$7.2 MB
\end{tabular}
}
\end{minipage}
\label{app:numparam_roberta}
\end{table}

\begin{table}[H]
\caption{The number of all parameters and those used for training and inference as well as the model size of methods for $\text{DistilBERT}_{base}$.}
\begin{minipage}[b]{1.0\hsize}\centering
\scalebox{0.80}{
\begin{tabular}{l | l | l | l | l }
           & \multicolumn{4}{c }{Fine-Tuning} \\  \hline
           & Trainable  & Inference  & Total   &  Total (Size)   \\ \hline \hline
Task = $\lbrace 1,10,30,60\rbrace$   & 66 M    &  66 M   &  66 M   &  264 MB   
\end{tabular}
}
\end{minipage}
\\
\begin{minipage}[b]{1.0\hsize}\centering
\vspace*{3mm}
\scalebox{0.82}{
\begin{tabular}{l | l | l | l | l }
           & \multicolumn{4}{c }{Adapters \& AdapterFusion}                       \\  \hline
           & Trainable  & Inference   & Total      & Total (Size)             \\ \hline \hline
Task = 1 \hspace*{2mm}  & 0.9 M       & 66.9 M     & 66.9 M   &  267.6 MB            \\ \hline
Task = 10  & 0.9 M  &  66 + (F$\times$0.9) M   & 75 M   & 300 MB                   \\ \hline
Task = 30  & 0.9 M  &  66 + (F$\times$0.9) M   & 93 M   & 372 MB                   \\ \hline
Task = 60  & 0.9 M  &  66 + (F$\times$0.9) M   & 120 M  & 480 MB
\end{tabular}
}
\end{minipage}
\\
\begin{minipage}[b]{1.0\hsize}\centering
\vspace*{3mm}
\scalebox{0.80}{
\begin{tabular}{l | l | l | l | l }
           & \multicolumn{4}{c }{ADA}                                       \\  \hline
           & Trainable  & Inference   & Total     & Total (Size)             \\ \hline \hline
Task = 1   & 0.9 M      &  66.9 M     &  66.9 M   & 267.6 MB   \\ \hline
Task = $\lbrace 10,30,60\rbrace$   &  2$\times$0.9 M  &  66.9 M   &  66 + (K+1)$\times$0.9 M  &  264 + (K+1)$\times$3.6 MB
\end{tabular}
}
\end{minipage}
\label{app:numparam_distilbert}
\end{table}

\begin{table}[H]
\caption{The number of all parameters and those used for training and inference as well as the model size of methods for ViT-B (Same for Deit-B).
$K$ is the number of Adapters in the pool, and $F$ is the number of fused Adapters (it is between 2 and number of tasks). For the \emph{Adapters} $F=1$.
For ER, for Task = $\lbrace 1,10,20\rbrace$, it is same with ADA Task=1. Total size is in MB.}
\begin{minipage}[b]{1.0\hsize}\centering
\scalebox{0.75}{
\begin{tabular}{l | l | l | l | l }
           & \multicolumn{4}{c }{Fine-Tuning (B1, B2) and EWC}  \\  \hline
           & Trainable  & Inference  & Total    &  Total (Size) \\ \hline \hline
Task = $\lbrace 1,10,20\rbrace$   & 86 M    &  86 M   &  86 M   &  344    
\end{tabular}
}
\end{minipage}
\\
\begin{minipage}[b]{1.0\hsize}\centering
\vspace*{3mm}
\scalebox{0.80}{
\begin{tabular}{l | l | l | l | l }
           & \multicolumn{4}{c }{Adapters \& AdapterFusion}                       \\  \hline
           & Trainable  & Inference   & Total          & Total (Size)             \\ \hline \hline
Task = 1 \hspace*{2mm}  & 1.8 M       &  87.8 M  &  87.8 M  & 351.2           \\ \hline
Task = 10  & 1.8 M  &  86 + (F$\times$1.8) M  &  104 M   & 416                 \\ \hline
Task = 20  & 1.8 M  &  86 + (F$\times$1.8) M  &  122 M   & 488                 
\end{tabular}
}
\end{minipage}
\\
\begin{minipage}[b]{1.0\hsize}\centering
\vspace*{3mm}
\scalebox{0.72}{
\begin{tabular}{l | l | l | l | l }
           & \multicolumn{4}{c }{ADA}                                       \\  \hline
           & Trainable  & Inference   & Total     & Total (Size)             \\ \hline \hline
Task = 1   & 1.8 M  & 87.8 M  & 87.8 M  & 351.2  \\ \hline
Task = 10  &  2$\times$1.8 M  & 87.8 M  & 86 + (K+1)$\times$1.8 M  &  344 + (K+1)$\times$7.2  \\ \hline
Task = 20  &  2$\times$1.8 M  & 87.8 M  & 86 + (K+1)$\times$1.8 M  &  344 + (K+1)$\times$7.2 
\end{tabular}
}
\end{minipage}
\label{table:numparam_vit}
\end{table}

Table~\ref{table:numparam_vit} reports the number of parameters used for baselines and ADA in image classification experiments with ViT and DeiT.
We don't add the head size to the table, since it's very small: 768 parameters per binary head, ~15K parameters (6 KB) for 20 tasks,
3840 per multi-class head, ~75K parameters (30KB) for 20 tasks. Also they are same for all the methods.

\begin{figure}[ht!]
\begin{minipage}{0.33\textwidth}
\centering
\includegraphics[width=\textwidth] {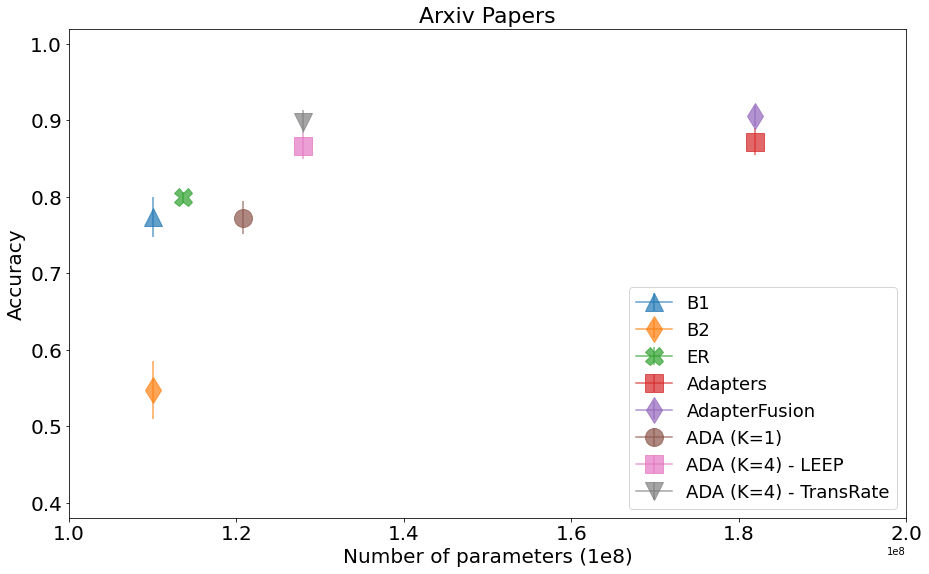}
\end{minipage}
\begin{minipage}{0.33\textwidth}
\centering
\includegraphics[width=\textwidth] {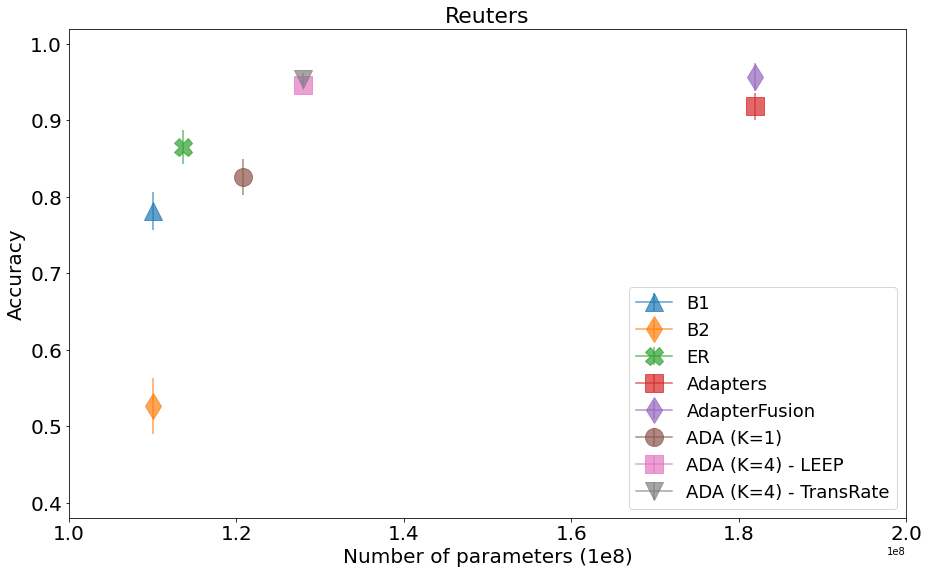}
\end{minipage}
\begin{minipage}{0.33\textwidth}
\centering
\includegraphics[width=\textwidth] {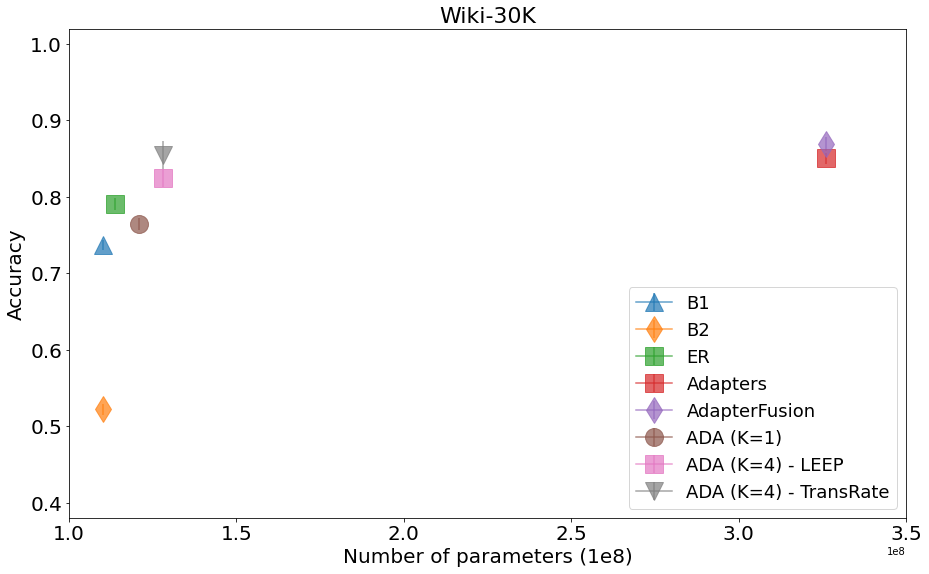}
\end{minipage}
\caption{Comparison of number of parameters of baselines and ADA on Arxiv, Reuters and Wiki-30K. 
The predictive performance reported on the y-axis is measured after processing all tasks.}
\label{fig:compMemory}
\vspace*{-2mm}
\end{figure}

Figure~\ref{fig:compMemory} and Table~\ref{app:numparamTable} show the number of parameters used by each method and their predictive performance.
These results make clear that \al{} is significantly more efficient in terms of memory usage. It can achieve 
predictive performance similar to the one of Adapters and AdapterFusion while requiring significantly less model parameters.
On Reuters and Arxiv, it can store the parameters of only 5 Adapters (K=4 Adapters in the pool, and one Adapter for new task),
against the 20 required by AdapterFusion.

\begin{table}[H]
\caption{The accuracy (with standard deviation) reported after last task and number of total parameters (Num Params) kept in memory for Adapters, AdapterFusion and ADA variants for experiments on Arxiv, Reuters, Wiki-30K and Mixed setting experiments with $\text{BERT}_{base}$.}
\begin{minipage}[b]{1.0\hsize}\centering
\vspace*{3mm}
\scalebox{0.75}{
\begin{tabular}{l | l | l | l | l | l | l | l | l}
                    &  \multicolumn{2}{c |}{Arxiv}       &   \multicolumn{2}{c |}{Reuters}   &   \multicolumn{2}{c |}{Wiki-30K}   &     \multicolumn{2}{c }{Mixed}    \\  \hline
                    & Accuracy & Num Params &  Accuracy  & Num Params      &  Accuracy   & Num Params  &  Accuracy  & Num Params        \\ \hline \hline
Adapters            &  0.872 $\pm$ 0.017    & 182 M   &  0.918 $\pm$ 0.015    &  182 M     &  0.847 $\pm$ 0.006   &   326 M   &  0.820 $\pm$ 0.008 &  470 M   \\ \hline
AdapterFusion       &  0.905 $\pm$ 0.013    & 182 M   &  0.957 $\pm$ 0.012    &  182 M     &  0.867 $\pm$ 0.009   &   326 M   &  0.833 $\pm$ 0.010 &  470 M   \\ \hline
ADA (K=1)           &  0.772 $\pm$ 0.021    & 120.8 M &  0.825 $\pm$ 0.018    &  120.8 M   &  0.770 $\pm$ 0.019   &  120.8 M  &  0.729 $\pm$ 0.017 &  120.8 M   \\ \hline
ADA (K=4) - LEEP    &  0.867 $\pm$ 0.017    & 128 M   &  0.947 $\pm$ 0.015    &  128 M     &  0.842 $\pm$ 0.013   &  128 M    &  0.795 $\pm$ 0.012 &  128 M     \\ \hline
ADA (K=4) -TansRate &  0.898 $\pm$ 0.014    & 128 M   &  0.951 $\pm$ 0.013    &  128 M     &  0.858 $\pm$ 0.011   &  128 M    &  0.812 $\pm$ 0.012 &  128 M    
\end{tabular}
}
\end{minipage}
\label{app:numparamTable}
\end{table}

\subsection{Inference and training time}
\label{a:inf_time}

\begin{figure}[h!]
     \centering
     \begin{subfigure}[b]{0.24\textwidth}
         \centering
         \includegraphics[width=28mm,height=32mm]{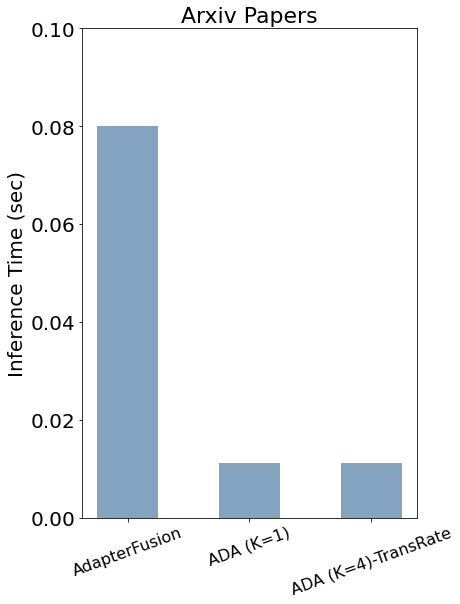}
         \caption{}
         \label{fig:arx_inf}
     \end{subfigure}
     \hfill
     \begin{subfigure}[b]{0.24\textwidth}
         \centering
         \includegraphics[width=28mm,height=32mm]{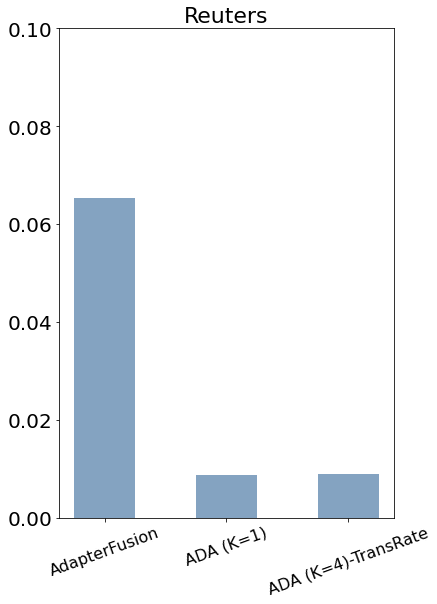}
         \caption{}
         \label{fig:reut_inf}
     \end{subfigure}
     \hfill
     \begin{subfigure}[b]{0.24\textwidth}
         \centering
         \includegraphics[width=28mm,height=32mm]{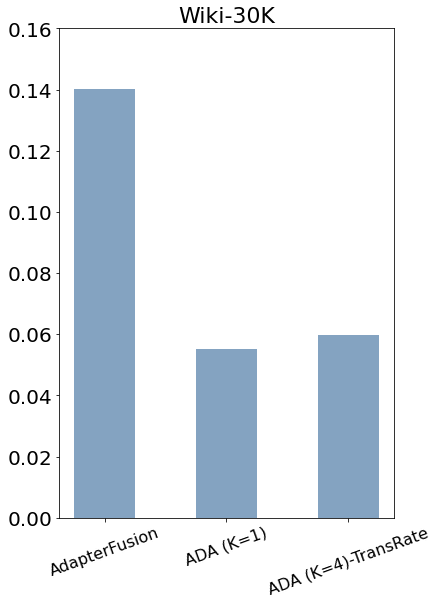}
         \caption{}
         \label{fig:wiki_inf}
     \end{subfigure}
     \hfill
     \begin{subfigure}[b]{0.24\textwidth}
         \centering
         \includegraphics[width=28mm,height=32mm]{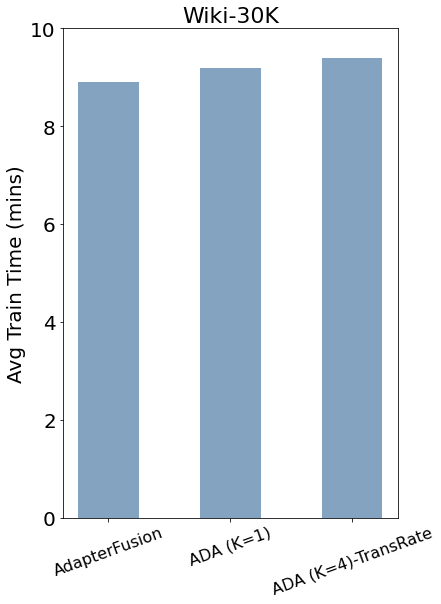}
         \caption{}
         \label{fig:wiki_train}
     \end{subfigure}
        \caption{Comparison of inference times of methods on a) Arxiv, b) Reuters and c) Wiki-30K. d) Comparison of training time on Wiki-30K.}
        \label{fig:compTime}
\end{figure}

In Figure~\ref{fig:arx_inf}, \ref{fig:reut_inf} and \ref{fig:wiki_inf} we report the average time per prediction made during our experiments.
We observe a significant speedup at inference time compared to AdapterFusion. For example, on Reuters, ADA is ~5 times faster than AdapterFusion
when both K=1 and K=4 (because it always uses one distilled Adapter for inference that has a fixed size).
The inference time of AdapterFusion depends on the number of Adapters fused.
Results demonstrate that \al{} provides a sufficiently fast inference for all datasets.

\textbf{Training time.} Distillation of Adapters brings an extra cost for ADA while learning fusion parameters brings an extra cost for AdapterFusion.
Computing transferability takes constant time which is negligible. Distillation costs training an additional Adapter (~1.6 \% of full fine-tuning time of BERT).
Figure~\ref{fig:wiki_train} reports the average training time comparison on Wiki-30K that is the largest difference with AdapterFusion
given larger number of tasks. We can clearly see that the difference is fractional while the difference between the inference time is significant.

\subsection{Memory consumption of ViT and DeiT}
\label{a:mem_vit_deit}

\begin{figure}[ht!]
\begin{minipage}{0.50\textwidth}
\centering
\includegraphics[scale = 0.20] {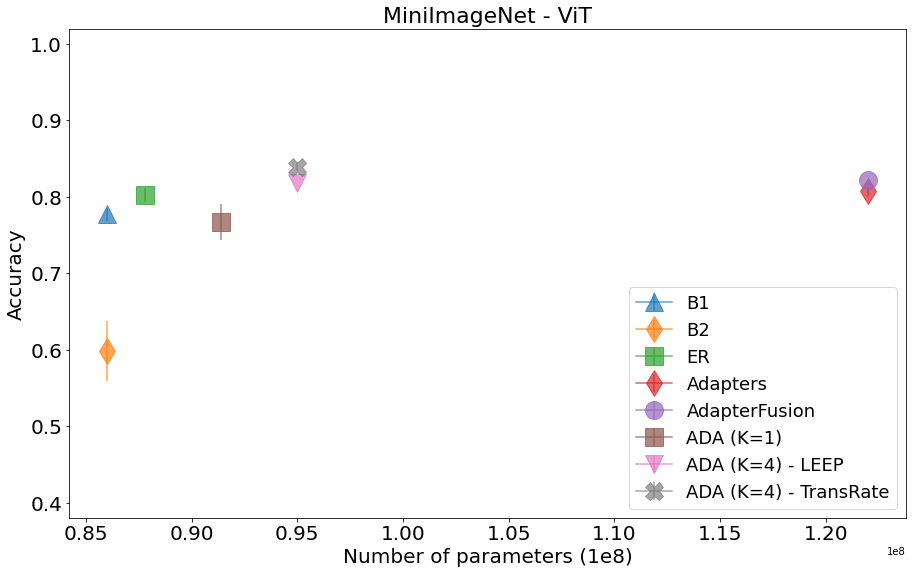}
\end{minipage}
\begin{minipage}{0.50\textwidth}
\centering
\includegraphics[scale = 0.20] {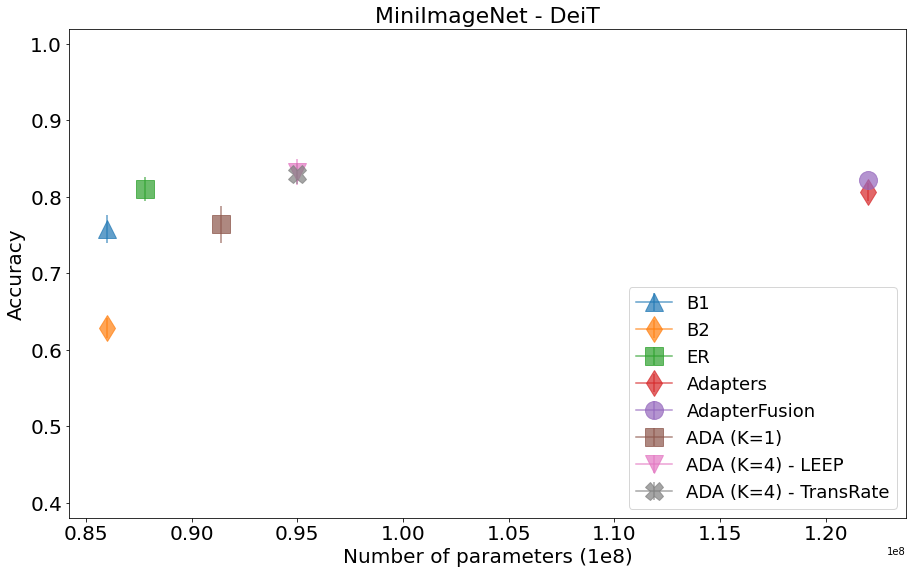}
\end{minipage}
\caption{Comparison of number of parameters of baselines and ADA on ImageNet with ViT and DeiT models. 
The predictive performance reported on the y-axis is measured after processing all tasks.}
\label{fig:compMemoryVitDeit}
\end{figure}

Figure~\ref{fig:compMemoryVitDeit} shows the number of parameters used by each method and their predictive performance. These results make clear that ADA is significantly 
more efficient in terms of memory usage also with ViT and DeiT models. It can achieve predictive performance similar to the one of Adapters and AdapterFusion
while requiring significantly less model parameters.

\subsection{Additional experiments with different task sizes}
\label{a:dataeff}

We would like to verify if the intelligent distillation mechanism we designed for \al{} is not only
able to avoid forgetting and save memory but also to increase the data efficiency.
Distilling together similar tasks for which a small number of data points is available could also provide a
better representation of the data points.

To verify this hypothesis, we repeated our experiments with a variable number of data points in the training set of each task.
The amount of positive and negative samples is balanced in both train and test tasks.
The size of the training sets of the Reuters tasks contain $t=\lbrace 20, 50, 80 \rbrace$ samples per class (positive and negative) 
and the test sets contain 20 samples per class. Arxiv Papers has more samples than Reuters dataset, so we added larger training tasks of size 400 to the configuration,
and increased the test task size. For Arxiv, we created the training sets with $t=\lbrace 20, 50, 100, 200 \rbrace$ samples per positive and negative classes
and the test set with 50 samples per class. 
Our expectation is that by increasing the training set size the overall predictive performance will improve, but we also
expect to see the predictive performance of \al{} matching (or narrowing the gap with) independent Adapters' one when using
a smaller training set. 

\begin{figure}[ht!]
\begin{minipage}{0.5\textwidth}
\centering
\includegraphics[scale = 0.2] {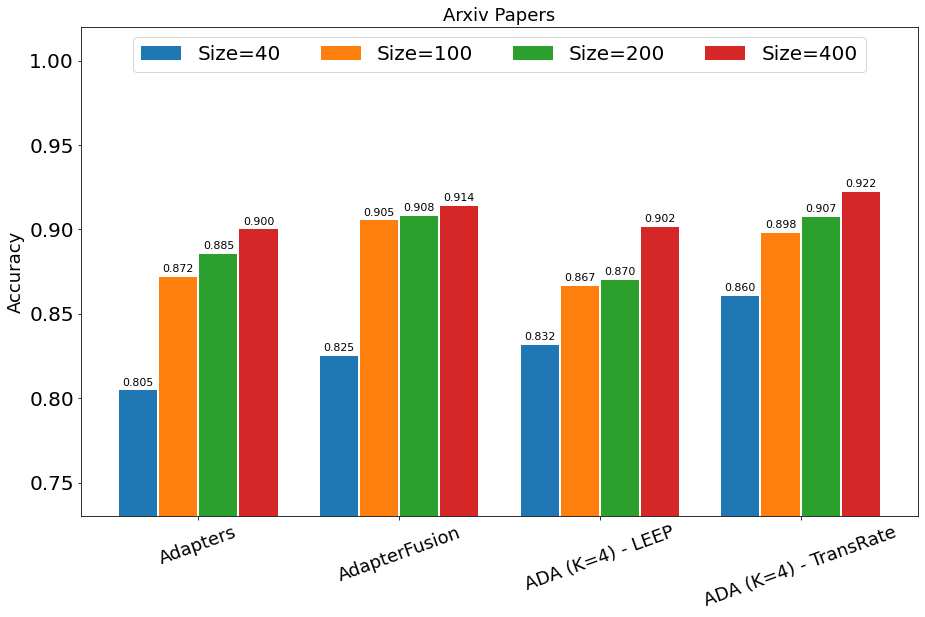}
\end{minipage}
\begin{minipage}{0.5\textwidth}
\centering
\includegraphics[scale = 0.2] {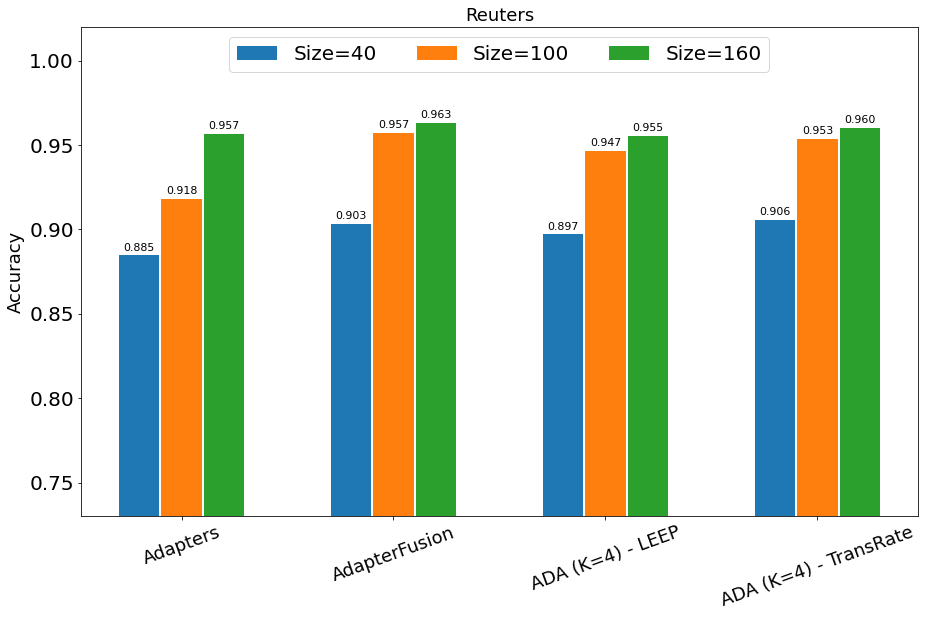}
\end{minipage}
\caption{Predictive performance of Adapter based methods with $t=\lbrace 20,50,100,200\rbrace$ on Arxiv and $t=\lbrace 20,50,80\rbrace$ on Reuters.}
\label{fig:ADA_numlabels}
\end{figure}

In Figure~\ref{fig:ADA_numlabels} we report the results of our experiment. We observe TransRate performing generally
better than LEEP, as in previous experiments. Focusing on TransRate, we can see that \al{} K=4 with TransRate can actually
outperform independent Adapters when the training set size is around 100 data points and even match the performance of 
independent Adapters using significantly more labels (200 labels on Arxiv and 160 on Reuters).
The effect becomes smaller or vanishes when the training set gets larger but this could still bring an important advantage
in the ``few-shot'' setting.

\subsection{Additional experiments with DistilBERT and Roberta}
\label{a:db_roberta}

We repeated all the experiments presented in Section~\ref{sec:text_classification} with $\text{DistilBERT}_{base}$ and $\text{RoBERTa}_{base}$ 
as our base models in order to show that it's not only limited to one specific model. The results demonstrated the same trends with 
$\text{BERT}_{base}$ model experiments. 

\begin{figure}[h!]
\begin{minipage}{0.5\textwidth}
\centering
\includegraphics[scale = 0.2] {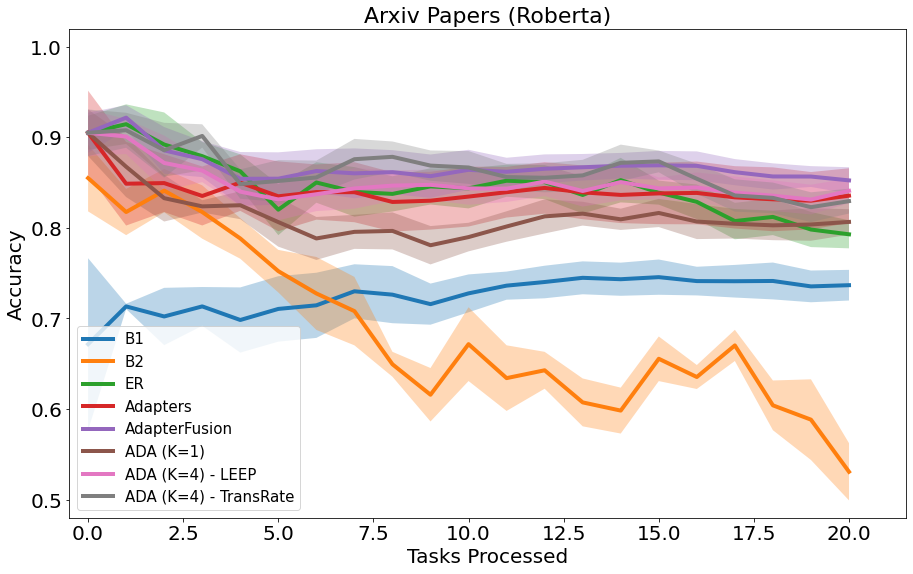}
\end{minipage}
\begin{minipage}{0.5\textwidth}
\centering
\includegraphics[scale = 0.2] {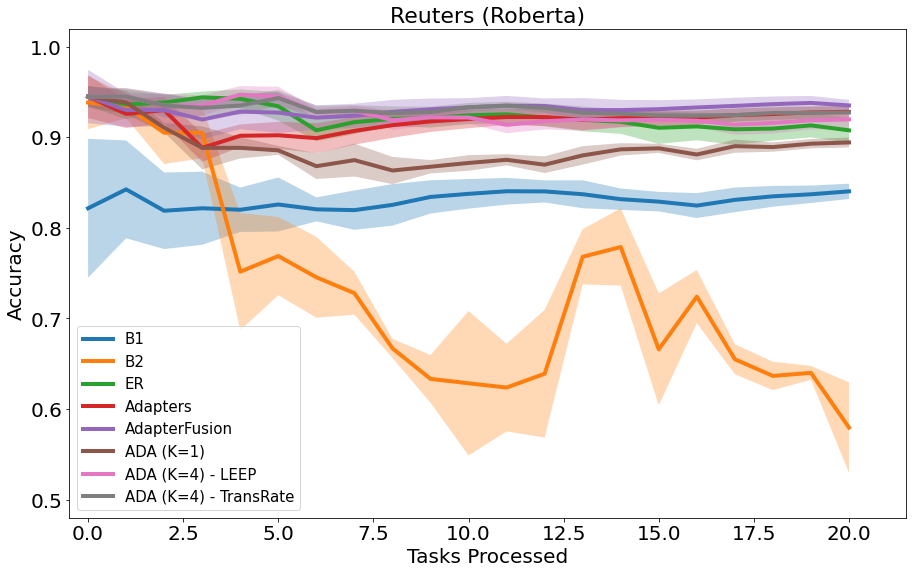}
\end{minipage}
\caption{Comparison of baselines and distillation methods on Arxiv and Reuters with $\text{RoBERTa}_{base}$. 
On the x-axis we report the number of tasks processed, on the y-axis we report the average accuracy measured 
on the test set of the tasks processed, shaded area shows standard deviation.}
\label{fig:compRobertaAccuracies}
\end{figure}

Figure~\ref{fig:compRobertaAccuracies} compares the ADA algorithms with baselines. The findings that we mention in predictive performance is
exactly applicable to $\text{RoBERTa}_{base}$ results. $\text{RoBERTa}_{base}$ performs slightly better on all the methods compared to $\text{BERT}_{base}$.
The behavior of algorithms are same for $\text{DistilBERT}_{base}$ and is very similar to the results with $\text{BERT}_{base}$, however, the number of parameters
used is different.

\begin{figure}[h!]
\begin{minipage}{0.5\textwidth}
\centering
\includegraphics[scale = 0.20] {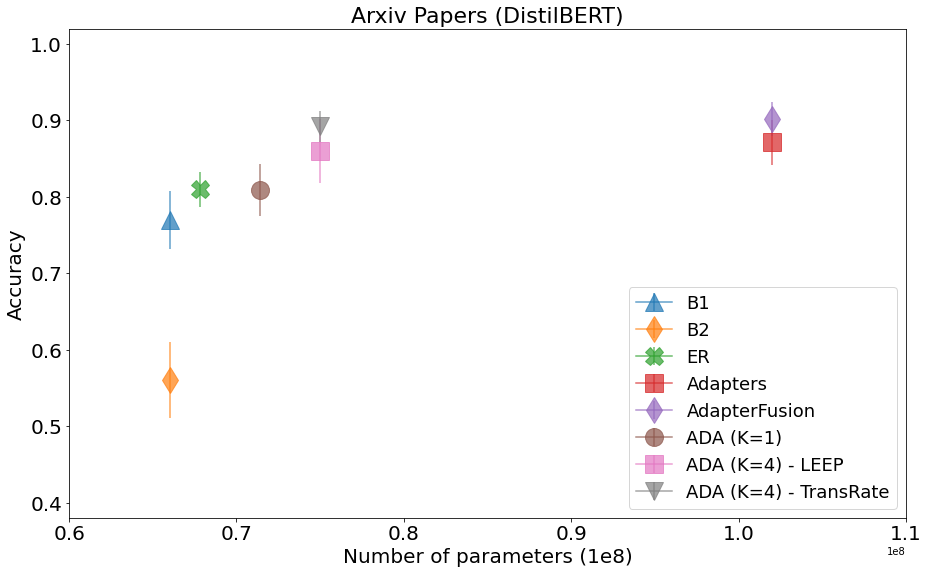}
\end{minipage}
\begin{minipage}{0.5\textwidth}
\centering
\includegraphics[scale = 0.20] {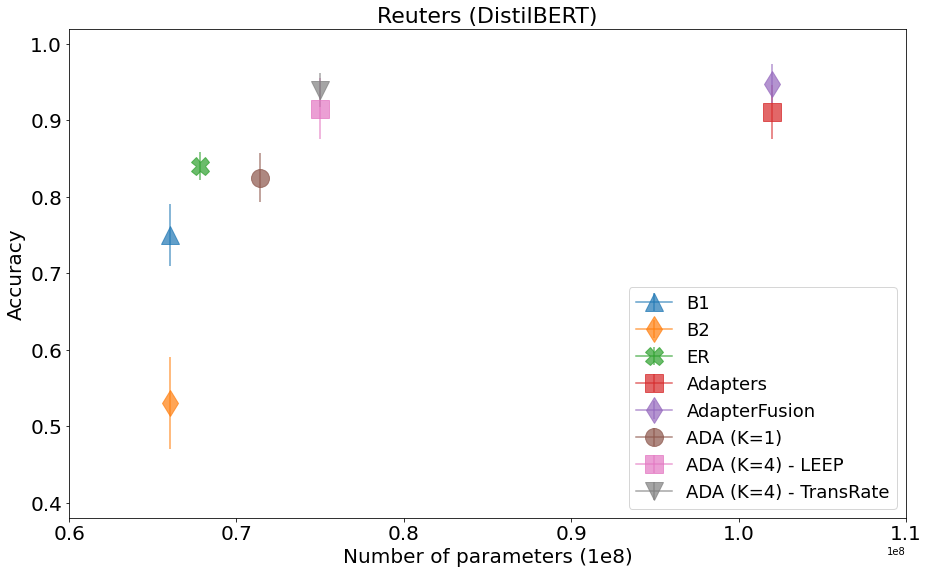}
\end{minipage}
\caption{Comparison of number of parameters of baselines and ADA on Arxiv, Reuters and Wikipedia with $\text{DistilBERT}_{base}$.}
\label{fig:compMemoryDB}
\end{figure}

Figure~\ref{fig:compMemoryDB} shows the number of parameters used by each method and their predictive performance with $\text{DistilBERT}_{base}$ model.
(We skip this figure for $\text{RoBERTa}_{base}$ because the number of parameters is very close to $\text{BERT}_{base}$, and we already show the 
accuracy in Figure~\ref{fig:compRobertaAccuracies}.) 

\begin{figure}[h!]
\begin{minipage}{0.5\textwidth}
\centering
\includegraphics[scale = 0.2] {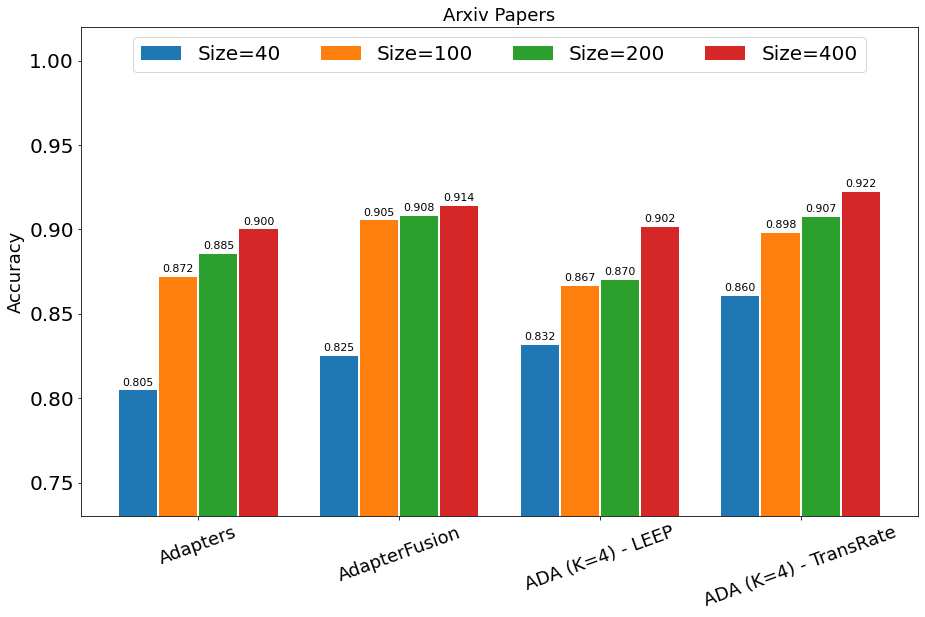}
\end{minipage}
\begin{minipage}{0.5\textwidth}
\centering
\includegraphics[scale = 0.2] {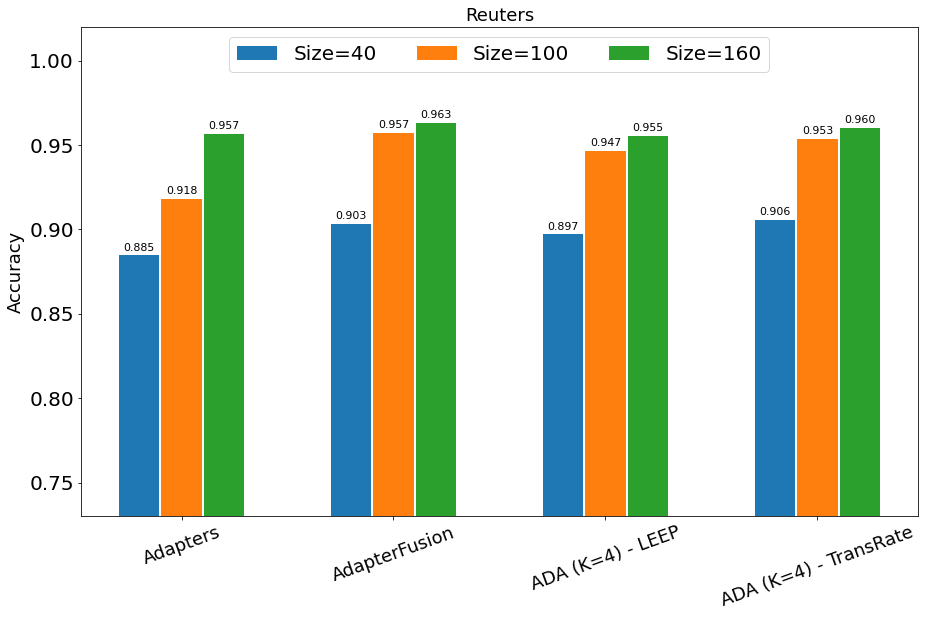}
\end{minipage}
\caption{Predictive performance of Adapter based methods with $t=\lbrace 20,50,100,200\rbrace$ on Arxiv and $t=\lbrace 20,50,80\rbrace$ on Reuters.}
\label{fig:ADA_numlabels_DB}
\end{figure}

As in \ref{a:dataeff}, we report the results of experiments of different task sizes with $\text{DistilBERT}_{base}$ in Figure~\ref{fig:ADA_numlabels_DB}.
This figure emphasises that with small number of labels and with a model much less parameters, we can still have good prediction accuracy on old and new tasks in CL setting.

\subsection{Additional experiments with DeiT}
\label{a:deit_exp}

We repeated all the experiments presented in Section~\ref{sec:image_classification} with DeiT-B~\cite{touvron2021training}
\footnote{\url{https://dl.fbaipublicfiles.com/deit/deit_base_patch16_224-b5f2ef4d.pth}} as our base model in order to show that it's not only 
limited to one specific model. The results demonstrated the same trends with ViT-B~\cite{dosovitskiy2020image}
\footnote{\url{https://huggingface.co/google/vit-base-patch16-224}} experiments. 

\begin{figure*}[ht!]
\centering
\begin{minipage}{0.45\textwidth}
\includegraphics[width=\textwidth]{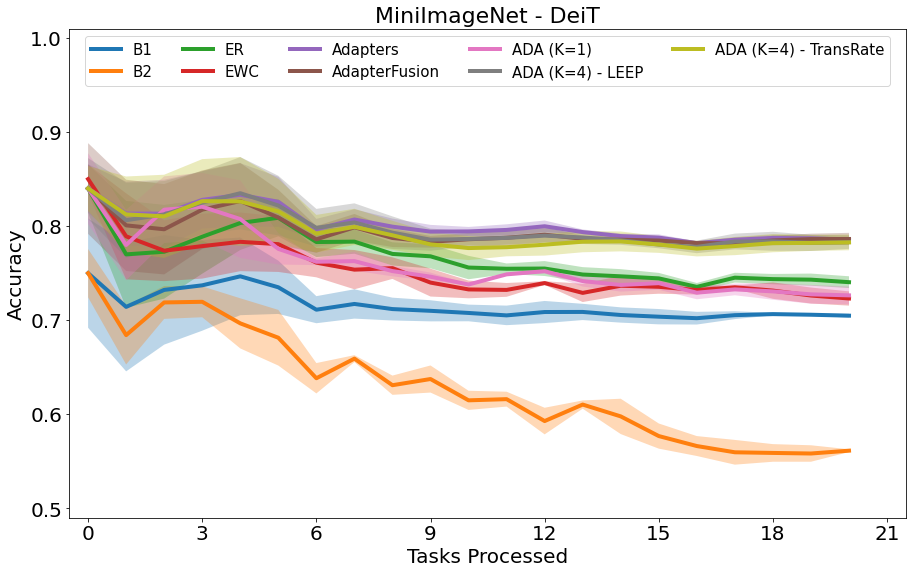}
\end{minipage}
\begin{minipage}{0.45\textwidth}
\includegraphics[width=\textwidth]{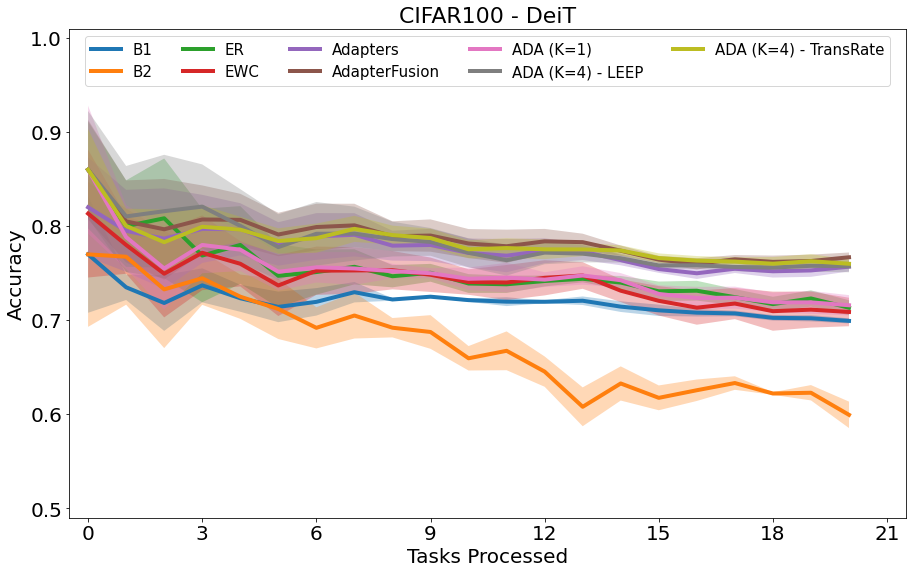}
\end{minipage}
\\
\centering
\begin{minipage}{0.45\textwidth}
\includegraphics[width=\textwidth]{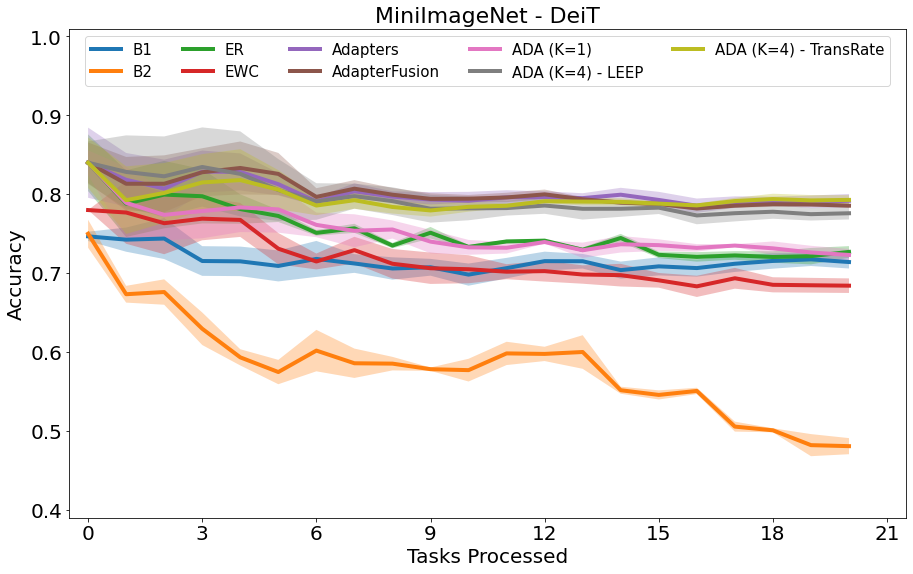}
\end{minipage}
\begin{minipage}{0.45\textwidth}
\includegraphics[width=\textwidth]{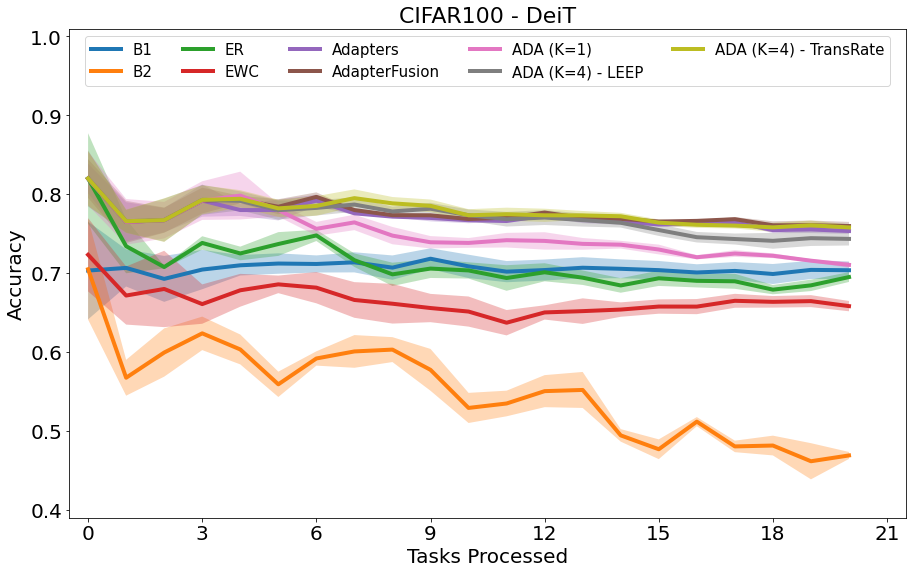}
\end{minipage}
\caption{Comparison between baselines and ADA with DeiT model on MiniImageNet and CIFAR100.
Top figures shows the first scenario (binary) results, and bottom figures shows the second scenario (multi-class) results.}
\label{fig:compAccuraciesDeiT}
\end{figure*}

\begin{figure*}[ht!]
\centering
\begin{minipage}{0.49\textwidth}
\includegraphics[width=\textwidth]{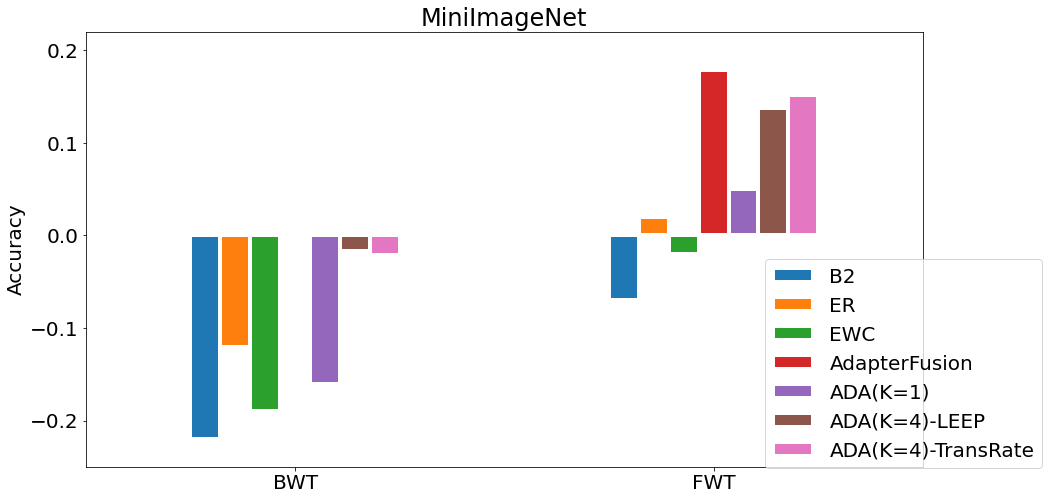}
\end{minipage}
\begin{minipage}{0.49\textwidth}
\includegraphics[width=\textwidth]{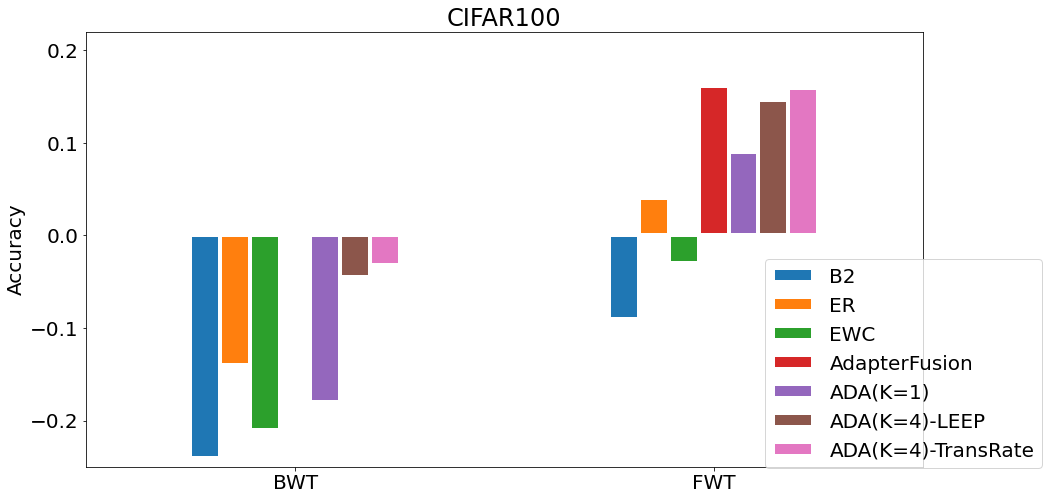}
\end{minipage}
\caption{Comparison between baselines and ADA with DeiT model on MiniImageNet and CIFAR100 for multi-class classification in terms of FWT and BWT.}
\label{fig:compFwtBwtDeiT}
\end{figure*}

In Figure~\ref{fig:compFwtBwtDeiT} we present FWT and BWT scores for baselines. As in Section~\ref{sec:text_classification}, we didn't present B1
and Adapters in the plots, since both FWT and BWT are zero for them. The behaviour is quite similar to text classification experiments.
BWT is zero for AdapterFusion, since the fusion parameter is computed with available Adapters, and the Adapters trained later is not used for
the previous tasks. ADA-LEEP and ADA-TransRate minimizes negative backward transfer, while showing a positive forward transfer for both
MiniImageNet and CIFAR100.

\newpage

\subsection{Additional experiments with different Adapters pool size}
\label{a:addpoolsize}

\begin{figure}[ht!]
\begin{minipage}{0.5\textwidth}
\centering
\includegraphics[scale = 0.20] {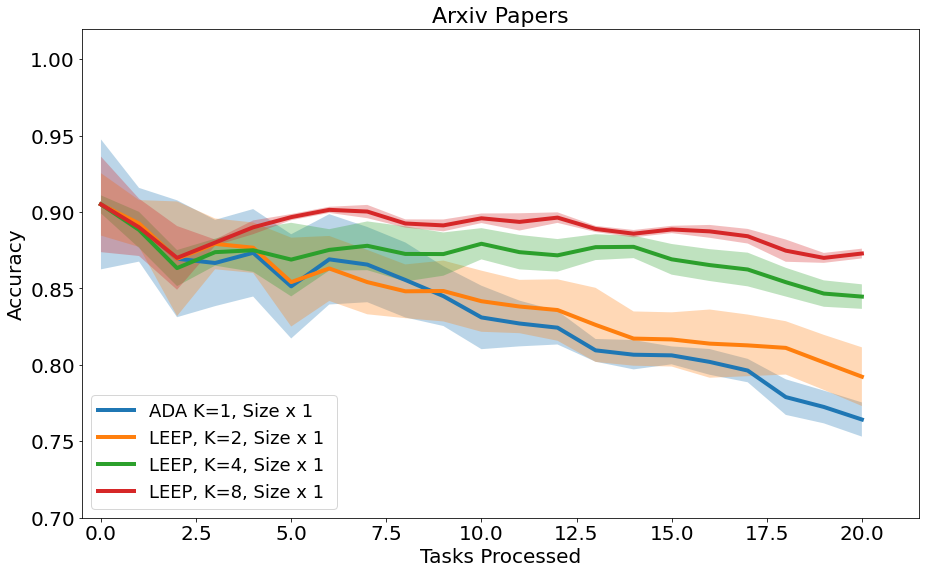}
\end{minipage}
\begin{minipage}{0.5\textwidth}
\centering
\includegraphics[scale = 0.20] {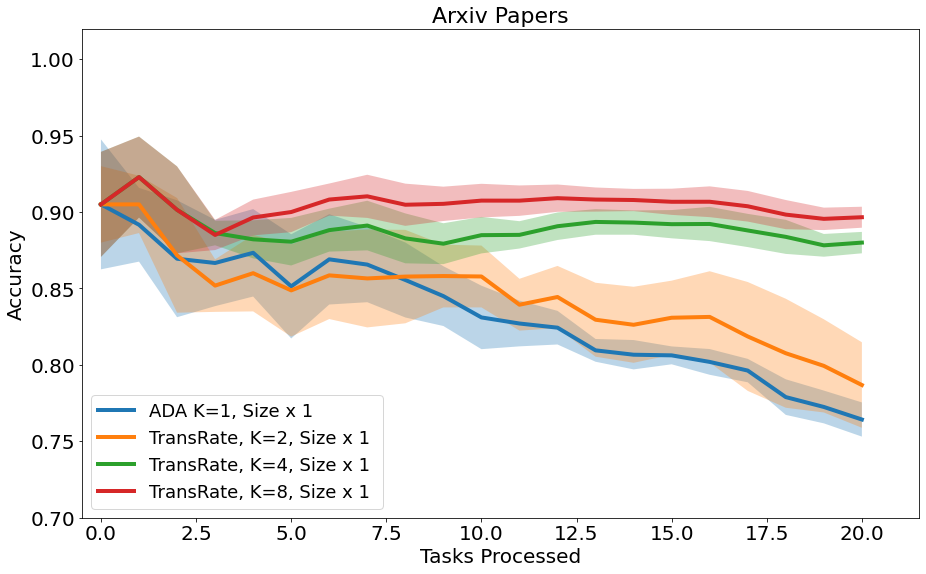}
\end{minipage}
\caption{Impact of Adapter pool size for \emph{LEEP} and \emph{TransRate} when $K=\lbrace 1,2,4,8 \rbrace$ on Arxiv for $t=50$.}
\label{a:poolsizearxiv2}
\end{figure}

This section has the additional results with different Adapters pool size on Arxiv Papers dataset.
As in Figure~\ref{fig:ADA_Kcomp}, the results in Figure~\ref{a:poolsizearxiv2} show a rapidly decreasing added value when the number of Adapters grows,
a behavior which aligns well with our practical requirements of keeping the number of model parameters under control when the number of tasks grows.

\end{document}